\documentclass[10pt,twocolumn,letterpaper]{article}

\usepackage[pagenumbers]{cvpr} 

\usepackage{graphicx}
\usepackage{amsmath}
\usepackage{amssymb}
\usepackage{booktabs}
\usepackage{adjustbox}

\usepackage[pagebackref,breaklinks,colorlinks]{hyperref}

\usepackage[capitalize]{cleveref}
\crefname{section}{Sec.}{Secs.}
\Crefname{section}{Section}{Sections}
\Crefname{table}{Table}{Tables}
\crefname{table}{Tab.}{Tabs.}

\def\cvprPaperID{4702} 
\def\confName{CVPR}
\def\confYear{2022}

\usepackage{overpic}
\usepackage{enumitem} 
\usepackage{color}
\usepackage{listings} 
\usepackage[skip=-.5em]{caption} 
\usepackage{xcolor}
\usepackage{currfile}
\usepackage{microtype} 
\usepackage{multirow}

\usepackage[rightcaption]{sidecap}  
\sidecaptionvpos{figure}{t} 
\definecolor{turquoise}{cmyk}{0.65,0,0.1,0.3}
\definecolor{purple}{rgb}{0.65,0,0.65}
\definecolor{dark_green}{rgb}{0, 0.5, 0}
\definecolor{orange}{rgb}{0.8, 0.6, 0.2}
\definecolor{red}{rgb}{0.8, 0.2, 0.2}
\definecolor{darkred}{rgb}{0.6, 0.1, 0.05}
\definecolor{blueish}{rgb}{0.0, 0.3, .6}
\definecolor{light_gray}{rgb}{0.7, 0.7, .7}
\definecolor{pink}{rgb}{1, 0, 1}
\definecolor{greyblue}{rgb}{0.25, 0.25, 1}

\definecolor{mydarkblue}{rgb}{0,0.08,0.55}
\hypersetup{  
    colorlinks=true,
    linkcolor=mydarkblue,
    citecolor=mydarkblue,
    filecolor=mydarkblue,
    urlcolor=mydarkblue
}

\newcommand{\CIRCLE}[1]{\raisebox{.5pt}{\footnotesize \textcircled{\raisebox{-.6pt}{#1}}}}

\newcommand{\loss}[1]{\mathcal{L}_\text{#1}}
\newcommand{\expect}{\mathbb{E}}

\newcommand{\Figure}[1]{Figure~\ref{fig:#1}}

\newcommand{\Table}[1]{Table~\ref{tab:#1}}

\newcommand{\Section}[1]{Section~\ref{sec:#1}}

\usepackage{blindtext}

\usepackage{lipsum}

\renewcommand{\paragraph}[1]{\vspace{.5em}\noindent\textbf{#1}.}

\newcommand{\RobustNeRF}{{Robust~NeRF}~}
\newcommand{\Kubric}{{Kubric}\xspace}

\definecolor{codegreen}{rgb}{0,0.6,0}
\definecolor{codegray}{rgb}{0.5,0.5,0.5}
\definecolor{codepurple}{rgb}{0.58,0,0.82}
\definecolor{backcolour}{rgb}{0.95,0.95,0.92}
\lstdefinestyle{mystyle}{
    backgroundcolor=\color{backcolour},   
    commentstyle=\color{codegreen},
    keywordstyle=\color{magenta},
    numberstyle=\tiny\color{codegray},
    stringstyle=\color{codepurple},
    basicstyle=\ttfamily\scriptsize,
    breakatwhitespace=false,   
    frame=tlb,framesep=4pt,framerule=0pt,
    breaklines=true,                 
    captionpos=b,                    
    keepspaces=true,                 
    numbers=left,                    
    numbersep=5pt,                  
    showspaces=false,                
    showstringspaces=false,
    showtabs=false,                  
    tabsize=2
}
\begin{document}
\title{Kubric: A scalable dataset generator}

\newcommand{\google}{1}
\newcommand{\uoft}{2}
\newcommand{\mcgill}{3}
\newcommand{\mila}{4}
\newcommand{\MIT}{5}
\newcommand{\deepmind}{6}
\newcommand{\ubc}{7}
\newcommand{\ucambridge}{8}
\newcommand{\eai}{9}
\newcommand{\haiper}{10}
\newcommand{\sfu}{11}

\author{
Klaus Greff\textsuperscript{\google} \and
Francois Belletti\textsuperscript{\google} \and
Lucas Beyer\textsuperscript{\google} \and
Carl Doersch\textsuperscript{\deepmind} \and
Yilun Du\textsuperscript{\MIT} \and
Daniel Duckworth\textsuperscript{\google} \and
David J Fleet\textsuperscript{\google,\uoft} \and
Dan Gnanapragasam\textsuperscript{\google} \and
Florian Golemo\textsuperscript{\mila, \eai} \and
Charles Herrmann\textsuperscript{\google} \and
Thomas Kipf\textsuperscript{\google} \and
Abhijit Kundu\textsuperscript{\google} \and
Dmitry Lagun\textsuperscript{\google} \and
Issam Laradji\textsuperscript{\mcgill, \eai} \and
Hsueh-Ti (Derek) Liu \textsuperscript{\uoft} \and
Henning Meyer\textsuperscript{\google} \and
Yishu Miao\textsuperscript{\haiper} \and
Derek Nowrouzezahrai\textsuperscript{\mcgill,\mila} \and
Cengiz Oztireli\textsuperscript{\google,\ucambridge} \and
Etienne Pot\textsuperscript{\google} \and
Noha Radwan\textsuperscript{\google} \and
Daniel Rebain\textsuperscript{\google,\ubc} \and
Sara Sabour\textsuperscript{\google,\uoft} \and
Mehdi S.\ M.\ Sajjadi\textsuperscript{\google} \and
Matan Sela\textsuperscript{\google} \and
Vincent Sitzmann\textsuperscript{\MIT} \and
Austin Stone\textsuperscript{\google} \and
Deqing Sun\textsuperscript{\google} \and
Suhani Vora\textsuperscript{\google} \and
Ziyu Wang\textsuperscript{\haiper} \and
Tianhao Wu\textsuperscript{\ucambridge} \and
Kwang Moo Yi\textsuperscript{\ubc} \and
Fangcheng Zhong\textsuperscript{\ucambridge} \and
Andrea Tagliasacchi\textsuperscript{\google,\uoft,\sfu} \and
\\
\textsuperscript{\google}Google Research \quad
\textsuperscript{\uoft}University of Toronto \quad
\textsuperscript{\mcgill}McGill University \quad
\textsuperscript{\mila}Mila \quad
\textsuperscript{\MIT}MIT \quad 
\textsuperscript{\deepmind}DeepMind \quad 
\\[.5em]
\textsuperscript{\ubc}UBC \quad
\textsuperscript{\ucambridge}University of Cambridge \quad
\textsuperscript{\eai}ServiceNow \quad
\textsuperscript{\haiper}Haiper \quad
\textsuperscript{\sfu}Simon Fraser University
}

\maketitle
\begin{abstract}
\vspace*{-0.2cm}
Data is the driving force of machine learning, with the amount and quality of training data often being more important for the performance of a system than architecture and training details.
But collecting, processing and annotating real data at scale is difficult,  expensive, and frequently raises additional privacy, fairness and legal concerns. 
Synthetic data is a powerful tool with the potential to address these shortcomings: 1) it is cheap 2) supports rich ground-truth annotations 3) offers full control over data and 4) can circumvent or mitigate problems regarding bias, privacy and licensing.
Unfortunately, software tools for effective data generation are less mature than those for architecture design and training, which leads to fragmented generation efforts.
To address these problems we introduce \emph{Kubric}, an open-source Python framework that interfaces with PyBullet and Blender to generate photo-realistic scenes, with rich annotations, and seamlessly scales to large jobs distributed over thousands of machines, and generating TBs of data.
We demonstrate the effectiveness of Kubric by presenting a series of 13 different generated datasets for tasks ranging from studying 3D NeRF models to optical flow estimation. 
We release Kubric, the used assets, all of the generation code, as well as the rendered datasets for reuse and modification.
\vspace{-1em}
\end{abstract}
\begin{figure}[t]
\begin{center}
\includegraphics[width=0.98\linewidth]{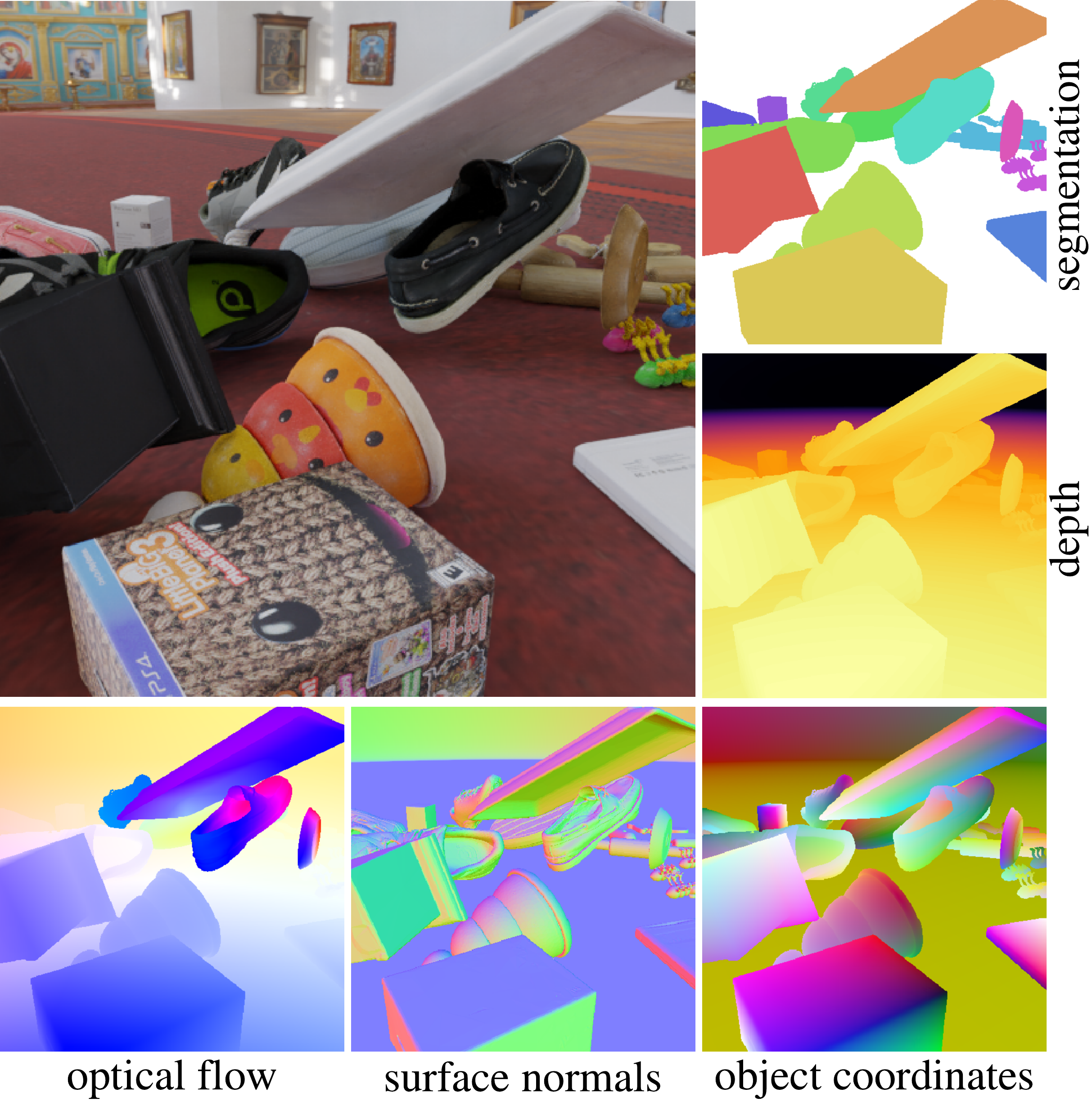}
\end{center}
\caption{
Example scene created and rendered with Kubric along with some of the automatically generated annotations.
}
\label{fig:teaser}
\vspace{-1em}
\end{figure}

\section{Introduction}
\label{sec:intro}
High quality data -- at scale -- is essential for deep learning.
It is arguably as or more important than many architectural and training details.
Nevertheless, even for many straightforward vision tasks, collecting and curating sufficient amounts of data continues to be a daunting challenge.
Some of the key barriers include the expense of high quality, detailed annotations, data diversity, control over task domain complexity, as well as concerns over privacy, fairness and licensing~\cite{asano21pass}.
This paper advocates the use of synthetic data to circumvent many of these problems, for which we introduce \Kubric, an open-source pipeline for generating realistic image and video data with rich ground truth annotations for myriad vision tasks.

Synthetic data has long been used for benchmark evaluation (e.g. for optical flow  \cite{Barron94performanceof,middleburyFlow11}), as it supports rich ground-truth annotations, and fine-grained control over data complexity. It also enables systematic model evaluation under violations of model assumptions (e.g. rigidity). 
Synthetic data has also been used effectively for training.
This includes seminal work on 3D human pose estimation from RGBD~\cite{Shotton-CVPR2011}, and more recently on myriad tasks, including facial landmark detection~\cite{wood2021fake}, human pose from video~\cite{Doersch-Zisserman-19}, and semantic segmentation~\cite{zhang21}.
Photo-realism is often considered essential for narrowing the generalization gap, but even without perfect realism, synthetic data can be remarkably effective (e.g., flying chairs~\cite{FlowNet-ICCV2015}, MPI-Sintel~\cite{ButlerECCV2012} and more recently AutoFlow~\cite{sun2021autoflow}).

Unfortunately, effective software tools for data generation are less mature than those for architecture design and training. It is therefore not surprising that  most generation efforts, although costly, have been one-off and task specific.
Although challenging to design and develop, what is needed is a
general framework for photo-realistic generation that supports reuse, replication, and shared assets, all at scale, enabling workflows with large jobs concurrently  on thousands of machines.
Kubric addresses these issues with coherent framework, a simple Python API, and a full set of tools for generation at scale, integrating assets from multiple sources, with a common export data format for porting data into training pipelines, and with rich annotations for myriad vision tasks.
%
%
In summary, our key contributions are:
\begin{itemize}
\itemsep 0in
\item We introduce \Kubric\footnote{\href{github.com/google-research/kubric}{https://github.com/google-research/kubric}}, a framework for generating photo-realistic synthetic  datasets for myriad vision tasks, with fine-grain control over data complexity and rich ground truth annotations.
\item Kubric enables generation at scale, seamlessly running large jobs over thousands of machines, generating TBs of data in a standard export data format.
\item The versatility of Kubric is demonstrated by the creation of 13 datasets for new vision challenge problems, spanning 3D NeRF models to optical flow estimation, along with benchmark results.
\end{itemize}

\section{Related Work}
\label{sec:related}
%
Synthetic data provides high quality labels for many image tasks such as semantic\cite{Chen_2019_CVPR} and instance\cite{ward2018deep} segmentation,  text localization\cite{Gupta_2016_CVPR}, object detection\cite{Hinterstoisser_2019_ICCV}, and classification\cite{frid2018synthetic}. There are many large synthetic datasets such as CLEVR\cite{johnson2017clevr}, ScanNet \cite{dai2017scannet}, SceneNet RGB-D~\cite{mccormac2017scenenet}, NYU v2 \cite{Silberman:ECCV12}, SYNTHIA~\cite{ros2016synthia}, virtual KITTI~\cite{gaidon2016virtual}, and flying things 3D~\cite{mayer2016large} for specific tasks. 
However, these datasets rarely contain all possible annotations for all image tasks, lacking key signals such as camera pose, instance or semantic segmentation masks, or optical flow. 
This is particularly challenging for multi-task problems like co-training a neural scene model with semantic segmentation\cite{zhi2021inplace}.
Moreover, fixed datasets can introduce biases\cite{tommasi2015deeper, torralba2011unbiased}  such as an object-centric bias~\cite{purushwalkam2020demystifying} and photographer's bias~\cite{azulay2019deep}.
By contrast, Kubric automatically generates the image cues for each frame and easily support a variety of viewing angles and lighting conditions.

\paragraph{Specialized Synthetic Data Pipelines}
There are many hand-crafted pipelines for synthetic data generation~\cite{nerf, Gupta_2016_CVPR,kim2021animeceleb} built off of rendering engines like Blender~\cite{blender} and Unity3D~\cite{borkman2021unity}.
While these pipelines mitigate biases in viewing angles and lighting, they are often specialized for a \textit{particular} task.  
This makes it challenging to adapt them to provide additional annotations without in-depth knowledge of the underlying rendering engine.
Real world to sim pipelines capture real-world data via 3D scans and then convert them into a synthetic data format.
\cite{OpenRooms2020} creates high quality room scenes, but has many manual steps including pose alignment and material assignment. 
\cite{eftekhar2021omnidata} also utilizes 3D scans, and provides control over a wide range of scene parameters, including camera position, field of view, and lighting, as well as a number of per frame image cues.
While these approaches produce high quality data for a particular captured scene, the pipeline still relies on 3D scans of the full scene, which imposes a bottleneck for scaling.


\def \y {$\checkmark$}
\def \n {$\times$}

\begin{table}[thb]
\footnotesize
    \centering
    \begin{tabular}{l|ccccc}
        Name                                        & Rendering & GI & Physics  & Scaling & DL \\
        \toprule
        Playing4Data~\cite{richter2016playing}      & (Game)    & \n & (Game)   & \n      & \n \\
        UnrealCV~\cite{qiu2016unrealcv}             & UE4       & \n & UE4      & \y      & \n  \\
        TDW~\cite{gan2020tdw}                       & Unity     & \n & PhysX    & \y      & \n  \\ %
        iGibson~\cite{xia2020interactive}           & PyRender  & \n & PyBullet & \y      & \n  \\
        Habitat~\cite{zot2021habitat}               & Magnum    & \n & Bullet   & \y      & \n  \\ %
        OpenRooms~\cite{OpenRooms2020}              & OptiX     & \y & --       & \n      & \n  \\
        Omnidata~\cite{eftekhar2021omnidata}        & Blender   & \y & --       & \n      & \y \\ %
        Blenderproc~\cite{denninger2019blenderproc} & Blender   & \y & Bullet   & \n      & \n \\ %
        Kubric                                      & Blender   & \y & PyBullet & \y      & \y \\
        \bottomrule
    \end{tabular}
    \vspace{1em}
    \caption{Rendering: Blender any OptiX are ray tracing engines, all others are based on rasterization; GI: support for global illumination; Physics: engine for physics simulation; Scaling: Easy to scale to very large datasets. DL: Data-loader integration with machine learning frameworks (PyTorch/TF).}
    \label{tab:comparison}
\end{table}

\newpage
\paragraph{Generic Dataset Creation Pipelines}
General purpose synthetic data pipelines (like Kubric) aim to address these issues by supporting arbitrary random compositions of meshes, textures, pre-existing scenes, etc. from collections of 3D assets. 
This mitigates some of the scaling considerations of real world to sim pipelines and more easily supports composition of assets from different datasets. 
These pipelines differ along various dimensions (see \cref{tab:comparison}). 
One important differences is the use of rendering engine, where ray-tracing engines support global illumination and other advanced lighting effects, which allow for a higher degree of realism than rasterization engines at the cost of a higher computational demand. 
Most general purpose synthetic data generation pipelines such as \cite{to2018ndds, kolve2019ai2thor, xia2020interactive, schwarz2020stillleben, DBLP:journals/corr/abs-1904-01201} are build on rasterization, which makes them very fast, often to the point of generating entire datasets on a single GPU machine.
ThreeDWorld~\cite{gan2020tdw} is an excellent example for such an engine with a flexible Python API, comprehensive export capabilities to a Unity3D based rasterization engine, the NVidia Flex~\cite{macklin2014flex} physics simulator and even sound generation via PyImpact~\cite{traer2019pyimpact}.
The framework closest in scope to Kubric is BlenderProc~\cite{denninger2019blenderproc}: a ray-tracing based pipeline built on Blender, which supports the generation of high-quality renders and comprehensive annotations as well as rigid-body physics simulations.
The main differences lie in Kubric's focus on scaling workloads to many workers, and its integration with 
tensorflow datasets.

\section{Infrastructure}

\begin{figure}
\includegraphics[width=\columnwidth]{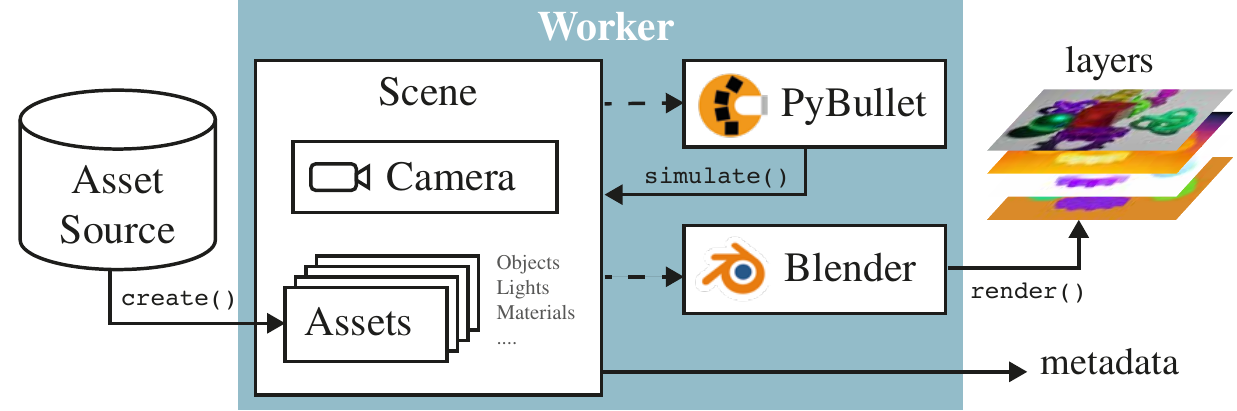}
\vspace{0em}  
\caption{\textit{Overview} -- a typical Kubric worker randomly populates a scene with assets loaded from an external source, possibly runs a physics simulation, renders the resulting frames, and finally exports the images, annotation layers, and other metadata. 
}
\label{fig:overview}
\vspace{-1.5em}
\end{figure}

Kubric is a high-level python library that acts as glue between: a rendering engine, a physics simulator, and data export infrastructure; see~\Figure{overview}.
Its main contribution is to streamline the process and reduce the hurdle and friction for researchers that want to generate and share synthetic data.

\subsection{Design Principles}
\paragraph{Openness} Data-generation code should be freely usable by researchers both in academia and in industry.
Kubric addresses this by being open-source with an Apache2 licence, and by only using software with similarly permissive licenses. 
Together with the use of free 3D assets and textures, this enables researchers to share not just the data, but also enable others to reproduce and modify it.

\paragraph{Ease of use} The fragmentation of computer graphics formats, conventions and interfaces is a major pain point for setting up and reusing data-generation code. 
Kubric minimizes this friction by offering a simple object-oriented API interface with PyBullet and Blender behind the scenes, hiding the complexities of setup, data transfer, and keeping them in sync.
We also provide pre-processed 3D assets from a variety of data sources, that can be used with minimal effort.

\paragraph{Realism} To be maximally useful, a data-generator should be able to model as much as possible of the structure and complexity of real data.
The Cycles raytracing engine of Blender supports a high level of realism and can model complex visual phenomena such as reflection, refraction, indirect illumination, subsurface-scattering, motion-blur, depth of field, etc.
Studying these effects is important, and they also help to reduce the generalization gap.

\paragraph{Scalability} Data generation workloads can range from simple toy-data prototyping all the way to generating massive amounts of high-resolution video data. 
To support this entire range of usecases, Kubric is designed to seamlessly scale from a local workflow to running large jobs on thousands of machines in the cloud. 

\paragraph{Portable and Reproducible} To facilitate reuse of data-generation code, it is important that the pipeline is easy to setup and produces the same results --- even when executed on different machines.
This is especially important due to the difficulty in installing the Blender Python module~\cite{blender2021bpy} and the substantial variations between versions.
By distributing a Kubric Docker image, we ensure portability and remove the bulk of the installation pain.

\paragraph{Data Export} Kubric by default exports a rich set of ground truth annotations from segmentation, optical flow, surface normals and depth maps, to object trajectories, collision events and camera parameters. 
We also introduce SunDs (see \cref{sec:sunds}), a unified multi-task frontend for richly annotated scene-based data.


\begin{figure*}
\noindent
\begin{minipage}[t]{.6\textwidth}
\begin{lstlisting}[language=Python]
import numpy as np
import kubric as kb
asset_source = kb.AssetSource("/assets/KuBasic")
scene = kb.Scene(resolution=(640, 640))
renderer = kb.renderer.Blender(scene)
simulator = kb.simulator.PyBullet(scene)
# --- populate the scene
scene.camera = kb.PerspectiveCamera(position=(0, 5, 5), 
                                    look_at=(0, 0, 0))
scene += kb.Cube(name="floor", scale=(10, 10, .1), 
                 position=(0, 0, -0.1), static=True)
scene += kb.PointLight(position=(-2.5, -1, 5), intensity=300)
rng = np.random.RandomState(seed=42)
for i in range(8): # place 8 random objects within a region
  mat = kb.PrincipledBSDFMaterial(color=kb.random_hue_color(rng=rng),
                                  metallic=rng.choice([0.0, 1.0]),
                                  transmission=rng.choice([0.0, 1.0]))
  obj = asset_source.create(rng.choice(asset_source.all_asset_ids),
                            material=mat, velocity=rng.normal(size=3))
  scene += obj
  kb.move_until_no_overlap(obj, simulator, rng=rng,
                           spawn_region=[[-1, -1, 0], [1, 1, 1]])
# --- execute the simulation, render, and save data to files
simulator.run()
renderer.save_state("output/scene.blend")
frames_dict = renderer.render()
kb.write_image_dict(frames_dict, "output")
kb.write_json({"camera": kb.get_camera_info(scene.camera),
               "instances": kb.get_instance_info(scene)},
              filename="output/metadata.json")
\end{lstlisting}
\end{minipage}
\hspace{1em}
\begin{minipage}[t]{.35\textwidth}
    \vspace{0.4em}
    \includegraphics[width=0.49\textwidth]{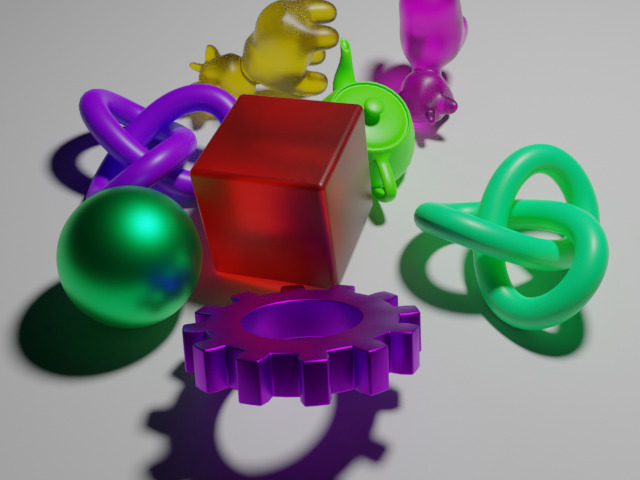}%
    \hspace{1pt}
    \includegraphics[width=0.49\textwidth]{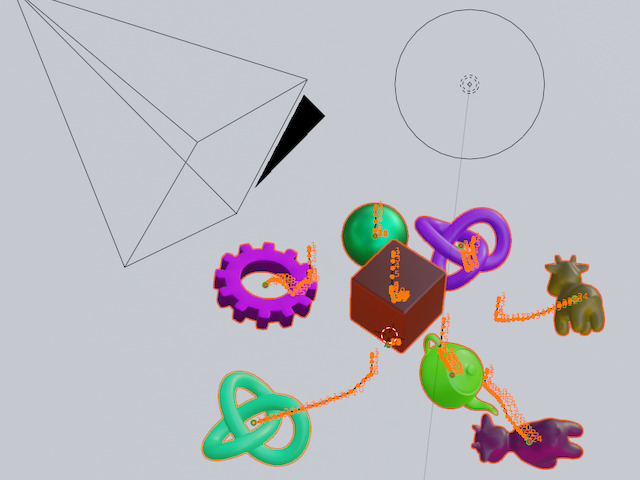}
    
    \vspace{0.2em}
    
    \includegraphics[width=0.25\textwidth]{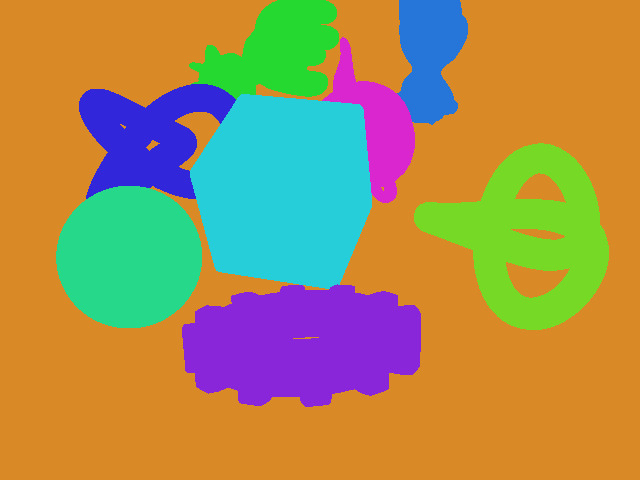}%
    \includegraphics[width=0.25\textwidth]{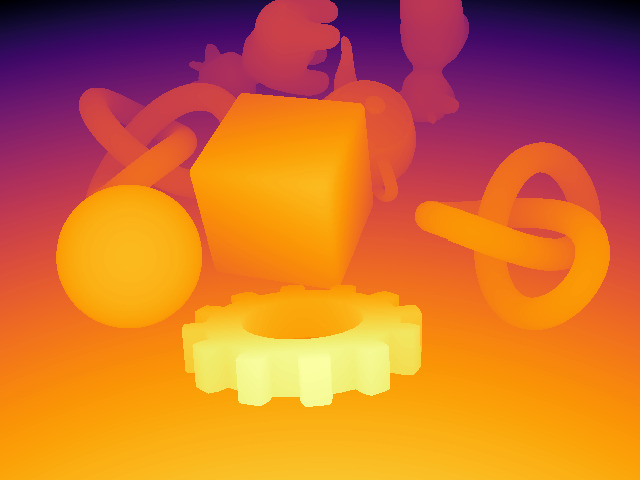}%
    \includegraphics[width=0.25\textwidth]{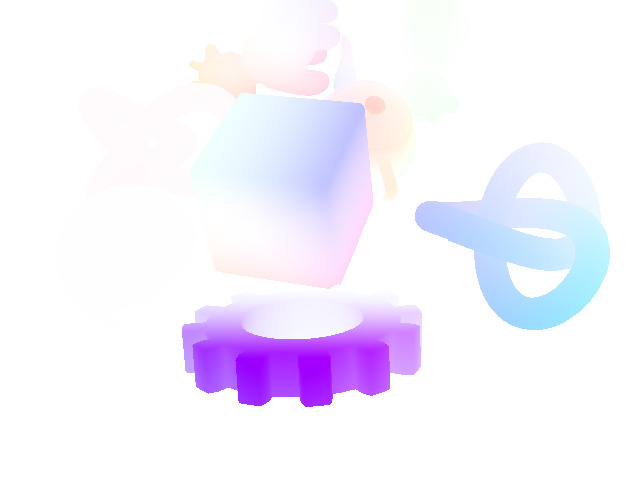}%
    \includegraphics[width=0.25\textwidth]{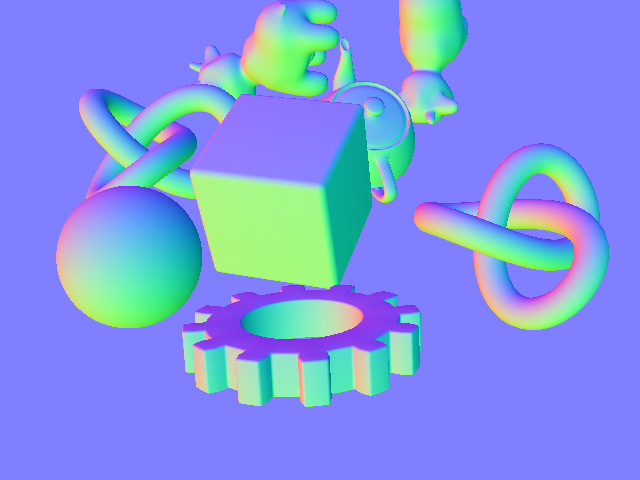}%
    
    \vspace{1em}
    
    \captionof{figure}{
    \textbf{Example worker} -- A simple environment with a floor, a point light, a perspective camera and eight KuBasic objects placed without overlap (by rejection sampling) and a random velocity.
    The scene physics is then simulated by the \texttt{PyBullet} backend, and rendered by the \texttt{Blender} backend.
    Infinite random variations of the scene can be generated by varying the random seed~(rng), and the result can be inspected inspected in Blender by opening the \texttt{.blend} file \textit{even before} rendering (top right).
    The exported image data includes annotations such as segmentation, depth, flow, and normals. 
    }
    \label{fig:code}
\end{minipage}
\vspace{-1em}
\end{figure*}

\subsection{Kubric Worker -- \Figure{code}}
The typical Kubric workflow consists of writing a worker script that creates, simulates, and renders a single random scene.
The full dataset is then generated by running this worker many times, and afterwards collecting the generated data.
This division into independent scenes mirrors the i.i.d. structure of most datasets and supports scaling the generation process from local prototyping to a large number of parallel jobs; e.g. using the Google Cloud Platform~(GCP), for which we provide convenience launcher scripts.
We also plan to support an Apache Beam pipeline that combines generation, collection and post-processing of datasets into a single convenient (but ultimately harder to debug) workflow.

\paragraph{Scene structure}
Each worker sets up a \texttt{Scene} object, which keeps track of global settings (e.g.,~ resolution, number of frames to render, gravity), a \texttt{Camera}, and all the objects, including lights, materials, animations, etc., which we refer to collectively as \texttt{Assets}.
They are the main abstractions used in Kubric to control the content of a scene, and they each expose a set of properties such as position, velocity or color.
When an \texttt{Asset} instance is added to the scene, the corresponding objects are created in each of the \texttt{Views}.
Currently this comprises the \texttt{PyBullet} simulator and the \texttt{Blender} renderer, but Kubric can be extended to support other views~(e.g.,~the recently open-sourced MuJoCo).
Kubric also maintains a link with the resulting data structure, and automatically communicates all changes to the properties of the assets to the connected views.
That way, the user only has to work with the abstractions provided by Kubric, and does not have to worry about differences in interfaces or conventions.

\paragraph{Simulator}
For physics simulation we interface with the open-source PyBullet physics engine~\cite{coumans2016pybullet} that is widely used in robotics (e.g., \cite{kalashnikov2018qtopt,xia2018gibson,james2019simtoreal}).
It can be used while populating the scene to ensure non-overlapping objects, but mainly it is used to run a (rigid-body) simulation, and to convert the resulting trajectories into keyframes and a list of collision events.
Bullet can also handle rigged models, softbody simulations, and various constraints that Kubric does not yet support.

\paragraph{Renderer}
Kubric uses the \texttt{bpy} module as an interface to Blender, a powerful open-source 3D computer graphics 
renderer which is widely used in game development and for visual effects.
Blender also comes with a powerful UI that can be used for interactively debugging and adjusting scenes, as well as creating and exporting new assets.
For rendering we rely on cycles -- Blender's raytracing engine -- which, unlike rasterized rendering engines, supports global illumination, accurately capturing effects such as soft shadows, reflection, refraction, and subsurface scattering.
These effects are crucial for visual realism,  and together with the vast set of other features of Blender, they enable artists to create photo-realistic 3D objects and scenes.
The downside is that cycles can be several orders of magnitude slower than a rasterized rendering engine, but for Kubric we decided that this computational cost is a worthwile tradeoff in exchange for the added realism and the ability to systematically study complex visual effects.

\paragraph{Annotations}
Another important feature of Blender is the use of specialized render passes that compute auxiliary ground truth information.
We leverage this feature to export (in addition to the RGB image) \CIRCLE{1} depth maps, \CIRCLE{2} instance segmentation, \CIRCLE{3} optical flow, \CIRCLE{4} surface normals, and \CIRCLE{5} object coordinates (see \cref{fig:teaser}).
In addition to these image space annotations, Kubric also automatically collects object-centric metadata, such as 2D/3D trajectories, 2D/3D bounding boxes, velocities, mass, friction, camera parameters, collision events, as well as custom metadata. 


\subsection{Assets}
A limiting factor in the creation of synthetic scenes is the availability of high-quality 3D assets. 
Several asset collections exist, but their use often requires substantial cleanup and conversion to make them compatible with a given pipeline. 
Kubric provides several preprocessed collections of assets in a public Google Cloud bucket.
Using these assets is as simple as changing the \texttt{path} of the asset source with~\texttt{kb.AssetSource(path)}.
At the core level, each dataset source is associated with a \texttt{manifest.json} file storing high level aggregated information, without the need to traverse the entire folder structure.
The \texttt{"id"} property of each entry in the manifest is in one-to-one correspondence to an archive file containing the data for the asset.
Each of these archives a contains a JSON-file with detailed metadata for a particular objects, including the paths to the sub-assets for rendering, collision detection, and the definition of physical properties in the Unified Robot Description Format (URDF) used by PyBullet.
For textured models, we employ the GLTF standard~\cite{gltf}, and store textures in binary format together with the geometry.

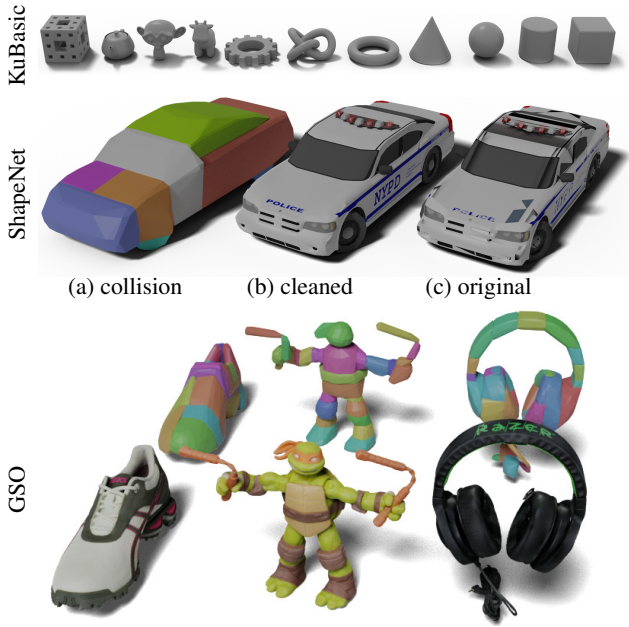
\begin{figure}
\hfill\begin{overpic}[width=0.95\linewidth]{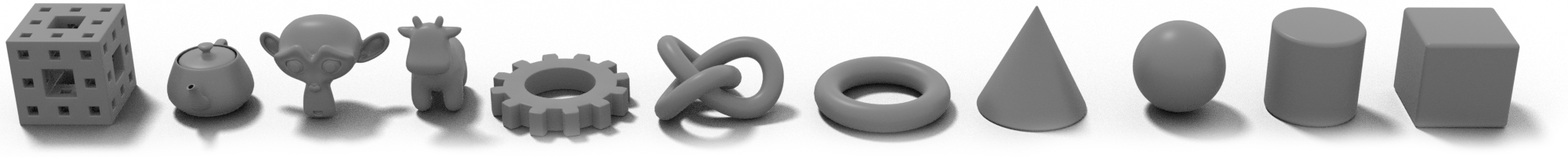}
\put(-5,-2){{\rotatebox{90}{\small{KuBasic}}}}
\end{overpic}

\vspace{1em}

\hfill\begin{overpic}[width=0.95\linewidth]{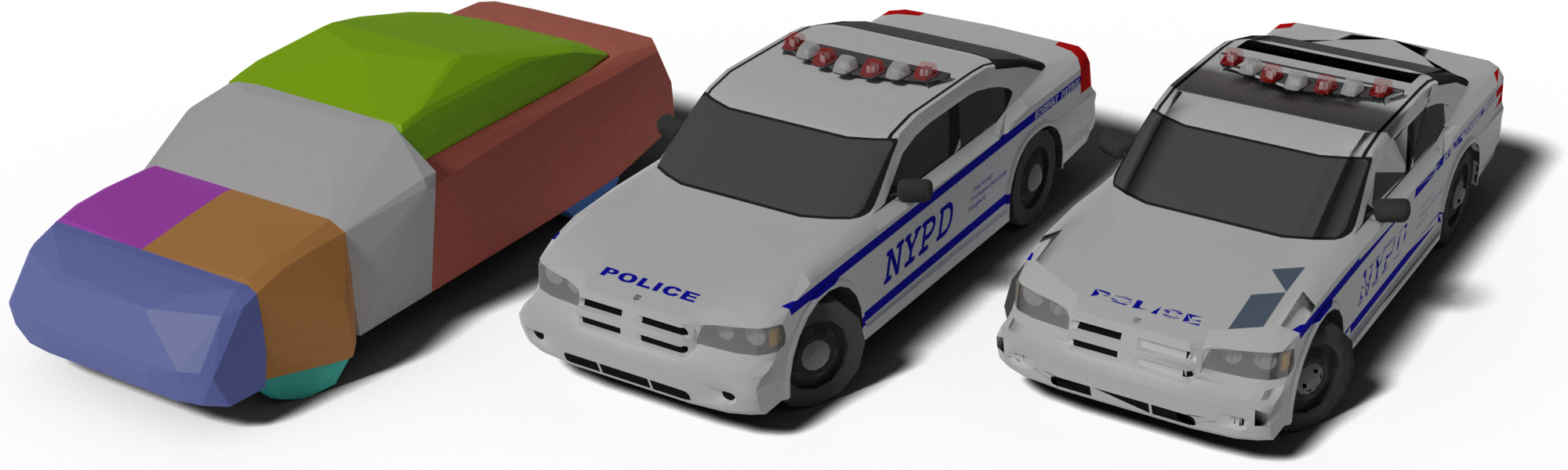}
  \put(-5,6){{\rotatebox{90}{\small{ShapeNet}}}}
  \put(5,-3){\small{(a) collision}}
  \put(35,-3){\small{(b) cleaned}}
  \put(65,-3){\small{(c) original}}
\end{overpic}

\vspace{1em}

\hfill\begin{overpic}[width=0.95\linewidth]{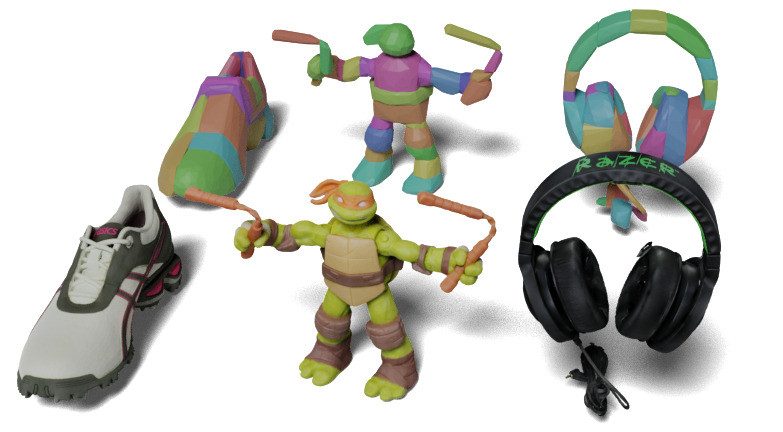}
  \put(-5,20){{\rotatebox{90}{\small{GSO}}}}
\end{overpic}

\vspace{1em}

\caption{(top) The KuBasic assets collection.
(middle)~ShapeNet objects by default do not render well in Blender (c) due to problems with auto-smoothing and the lack of backface culling in cycles. 
(b) We processed all ShapeNet objects to fix these issues and (a) generated a collision mesh by first making the model watertight and then performing an approximate convex decomposition using~VHACD.
(bottom)~Example assets from the Google Scanned Objects~(GSO) dataset along with the generated collision meshes.}
\label{fig:kubasic}
\label{fig:shapenet_convert}
 \label{fig:gso}
\vspace{-1em}
\end{figure}
\paragraph{KuBasic}
For simple prototyping we ship a small collection of eleven simple assets depicted in the top row of~\cref{fig:kubasic}. 

\paragraph{ShapeNetCore.v2}
This dataset is a subset of the full ShapeNet dataset~\cite{chang2015shapenet} with~$51,300$ unique 3D models from~$55$ with canonical alignment and common object categories annotations~(both manually verified).
Extensive pre-processing was performed to simplify the integration of these assets within \Kubric. These conversion scripts are available in the \texttt{shapenet2kubric} sub-directory; the conversion process can be easily reproduced by the Docker container available therein. 
This conversion process took $\approx16$ hours on a 80 core virtual machine~(Google Cloud VM~\texttt{n2-highcpu-80}) and parallelization across threads executed by Python's \texttt{multiprocessing} library.
A small set of models failed to convert (e.g., they had missing textures, erroneous materials, or simply crashed the conversion process) and are listed in the conversion code.

While many of the models within the dataset produce satisfactory renderings when visualized through OpenGL,  rendering quality is significantly higher when a \textit{photorealistic} renderer is employed~(i.e., Blender Cycles).
We collected the community's wisdom (i.e., ShapeNet and Blender official forums) on how to tweaks models to minimize the occurrence of visual artifacts.
The conversion procedure is automated via scripted Blender modifiers and involves removing doubles, disabling auto-smoothing, splitting sharp edges, and infinitesimally displacing the faces of polygonal meshes along the primitive's local normal.
For the collision geometry, we first converted the assets into watertight meshes with ManifoldPlus~\cite{manifoldplus}, and then resorted to the VAHCD~\cite{vhacd} implementation wrapped within PyBullet~\cite{pybullet} to compute the convex decomposition of a 3D object, whose mass and inertia tensors were finally estimated by trimesh~\cite{trimesh}.

\paragraph{Google Scanned Objects (GSO)~\cite{google2021gso}}
Is a dataset of common household objects that have been 3D scanned for use in robotic simulation and synthetic perception research. 
It is licensed under the CC-BY 4.0 License and contains $\approx1$k high-quality textured meshes; see~\cref{fig:gso}. 
We publish pre-processed version of this dataset in the Kubric format, which again includes generated collision meshes.

\paragraph{Polyhaven~\cite{polyhaven2021}} is a public (CC0 licensed) library from which we have collected and pre-processed HDRI images for use as backgrounds and lighting, and textures for use in high-quality materials. 


\subsection{Scene Understanding Datasets~(SunDs)}
\label{sec:sunds}
To facilitate ingesting data into machine learning models, we introduce, alongside Kubric, the SunDs (Scene Understanding Datasets) dataset front-end\footnote{\href{github.com/google-research/sunds}{https://github.com/google-research/sunds}}.
SunDs is an API to access public scene understanding datasets.
The field names and structure, shape, dtype are standardized across datasets. This allow to trivially switch between datasets (e.g. switch from synthetic to real data).
All SunDs datasets are composed of two sub-datasets:
\begin{itemize}
\setlength\itemsep{0em}
\item The scenes dataset contains high level scene metadata (e.g., scene boundaries, mesh of the full scene, etc.).
\item The frames dataset contains the individual examples within a scene (e.g., RGB image, bounding boxes, etc.).
\end{itemize}
SunDs abstracts away the dataset-specific file format~(json, npz, folder structure, \dots), and returns tensors directly ingestible by machine learning models (TF, Jax, Torch).
Internally, SunDs is a wrapper around TFDS, which allows one to scale to huge datasets ($\approx$~TB), to provide native compatibility with distributed cloud file systems~(e.g. GCS, S3), and to leverage \texttt{tf.data} pipeline capabilities~(prefetching, multi-threading, auto-caching, transformations, etc.).

To simplify even further data ingestion, SunDs introduce, on top of TFDS, the concept of tasks. Each SunDs dataset can be loaded for different tasks. Tasks control:
\begin{itemize}
\setlength\itemsep{0em}
    \item Which features of the dataset to use/decode. Indeed,  scene understanding datasets often have many fields (lidar, optical flow, \dots), but only a small subset are used for any given task. Selecting which fields are used avoids the cost of decoding unnecessary features.
    \item Which transformation to apply to the pipeline. For example, the NeRF task will dynamically generate the rays origin/directions from the camera intrinsics/extrinsics contained in the dataset.
\end{itemize}


\section{Kubric Datasets and Challenges}
\label{sec:challenges}

\def \y {$\checkmark$}
\def \n {$\times$}
\newcommand{\rot}[2][l]{\rotatebox[origin=#1]{90}{#2}}  

\begin{table}[t]
\centering
\setlength{\tabcolsep}{3pt}
\def\arraystretch{1}
\resizebox{\linewidth}{!}{ 
\begin{tabular}{clc|ccccc|ccccc|cccc|c}
   &
  Section &
  \rot{task domain} & 
  \rot{flow} &
  \rot{segmentation} &
  \rot{depth} &
  \rot{camera 3D pose} &
  \rot{object 3D poses} &
  \rot{physics sim.} &
  \rot{rigged animation} &
  \rot{control backgrnd.} &
  \rot{control materials} &
  \rot{control lighting} &
  \rot{new challenge} &
  \rot{sim-to-real} &
  \rot{hypothesis testing} &
  \rot{PII / legal} &
  \rot{scale}
  \\
\midrule
\ref{sec:object_discovery}   & {Object discovery}     & 2D &  \y & \y & \n & \n & \n &   \y & \n & \y & \y & \y &    \y & \n & \y & \n & \textit{TB} \\
\ref{sec:optical_flow}       & {Optical flow}         & 2D &  \y & \n & \n & \n & \n &   \y & \n & \y & \n & \y &    \n & \y & \n & \n & \textit{TB} \\
\ref{sec:texture_structure}  & {NeRF \& Texture}      & 3D &  \n & \n & \y & \n & \n &   \n & \n & \n & \y & \n &    \n & \n & \y & \n & \textit{MB} \\
\ref{sec:pose_estimation}    & {Pose-estimation}      & 2D &  \n & \n & \n & \n & \y &   \n & \y & \y & \y & \y &    \n & \y & \n & \y & \textit{GB} \\
\ref{sec:pretraining}        & {Pre-training}         & 2D &  \n & \n & \n & \n & \n &   \n & \n & \n & \y & \y &    \n & \y & \n & \y & \textit{GB} \\
\ref{sec:robust_nerf}        & {Robust NeRF}          & 3D &  \n & \n & \n & \y & \n &   \n & \n & \n & \y & \n &    \y & \n & \y & \n & \textit{MB} \\
\ref{sec:multi_view_sod}     & {Multi-view SOD}       & 2D &  \n & \y & \n & \n & \n &   \n & \n & \y & \n & \y &    \n & \y & \y & \n & \textit{GB}\\
\ref{sec:complex_brdfs}      & {Complex BRDFs}        & 3D &  \n & \n & \n & \y & \n &   \n & \n & \n & \y & \y &    \y & \n & \y & \n & \textit{GB} \\
\ref{sec:recons3d}           & {3D reconstruction}    & 3D &  \n & \y & \n & \y & \y &   \n & \n & \n & \n & \n &    \y & \n & \n & \n & \textit{GB} \\
\ref{sec:robust3d_recons}    & {Robust 3D recons.}    & 3D &  \y & \y & \n & \n & \y &   \n & \y & \y & \y & \y &    \y & \n & \y & \y & \textit{MB} \\
\ref{sec:point_tracking}     & {Point tracking}       & 2D &  \n & \y & \y & \y & \y &   \y & \n & \y & \n & \y &    \y & \n & \n & \n & \textit{TB} \\
\ref{sec:toybox}             & {ToyBox}               & 3D &  \n & \y & \y & \y & \y &   \n & \n & \y & \y & \y &    \y & \n & \y & \n & \textit{GB} \\
\ref{sec:nvs}             & {Novel View Synthesis} & 3D &  \n & \n & \n & \y & \n &   \y & \n & \y & \n & \y &    \y & \n & \y & \n & \textit{GB} \\
\bottomrule
\end{tabular}
} 
\vspace{1em}
\caption{
Overview of the datasets / challenges presented in \cref{sec:challenges}.
}
\label{tab:\currfilebase}
\vspace{-1em}
\end{table}

To demonstrate the power and versatility of Kubric, we next describe a series of new  challenge problems, each with data\footnote{The presented datasets along with the corresponding worker scripts can be found at \href{https://github.com/google-research/kubric}{https://github.com/google-research/kubric}.} generated by Kubric (see \cref{tab:challenges}).
They cover 2D and 3D tasks at different scales, with dataset sizes ranging from MBs to TBs.
Each relies on a different subset of annotations (flow, segmentation, depth, camera pose, or object pose), makes use of a different subset of features (e.g., physics or rigged animation), and
requires control over different factors (background, materials, or lighting).
Any one dataset might have been generated by a simpler, specialized code-base, but this would have been extremely inefficient.  Rather, with the versatility of Kubric, it was straightforward to create,
extend and combine datasets, leveraging 
a common platform and shared engineering efforts.

These different challenges also highlight different uses of synthetic data. 
Some serve as benchmarks for comparing existing and future methods, while others provide additional training data for real-world applications (sim-to-real).
Some are designed to empirically test specific hypotheses (e.g., in testing), while some focus on data that can be shared without privacy and legal concerns.

We describe 13 challenges in sections below; i.e., 
\ref{sec:robust_nerf}) Nerf reconstruction in the presence of non-static scenes structure;
\ref{sec:multi_view_sod}) a novel multi-view version of the salient object detection task;
\ref{sec:complex_brdfs}) a 3D reconstruction benchmark focussed non-Lambertian surfaces;
\ref{sec:recons3d}) a study on scaling and generalization of \textsc{SoftRas} for reconstructing 3D meshes from a single image;
\ref{sec:robust3d_recons}) a study on the robustness of video-based reconstruction of 3D meshes with respect to mesh deformations and inaccurate flow; 
\ref{sec:point_tracking}) a long-term dense 2D point tracking task including a novel contrastive tracking algorithm;
\ref{sec:toybox}) a large-scale multi-view semantic scene understanding benchmark; and
\ref{sec:nvs}) a challenging new-scene novel view synthesis dataset.

\begin{figure}[t]

\resizebox{\linewidth}{!}{ 
\begin{tabular}{@{}lcccccc@{}}
\toprule
Method & MOVi-A & MOVi-B & MOVi-C & MOVi-D & MOVi-E \\
\midrule
SAVi~\cite{anonymous2022conditional} & $\mathbf{82.0} \scriptstyle{\,\pm\,0.3}$ & $\mathbf{61.5} \scriptstyle{\,\pm\,0.3}$ & $\mathbf{47.0} \scriptstyle{\,\pm\,0.3}$ & $19.4 \scriptstyle{\,\pm\,8.0}$ & $2.7 \scriptstyle{\,\pm\,0.5}$
 \\
SIMONe~\cite{kabra2021simone} &  $61.8 \scriptstyle{\,\pm\,2.0}$ &  $30.7 \scriptstyle{\,\pm\,3.3}$ &  $19.8 \scriptstyle{\,\pm\,0.5}$ & $\mathbf{34.1} \scriptstyle{\,\pm\,0.7}$  & $\mathbf{34.9} \scriptstyle{\,\pm\,0.6}$ \\
\midrule
SAVi + BBox & $95.3 \scriptstyle{\,\pm\,0.2}$ & $85.5 \scriptstyle{\,\pm\,0.2}$ & $73.5 \scriptstyle{\,\pm\,0.3}$ & $9.9 \scriptstyle{\,\pm\,1.4}$ & $7.5 \scriptstyle{\,\pm\,1.0}$ \\     
\bottomrule
\end{tabular}
} 
\vspace{1em}
\captionof{table}{\textbf{Object discovery --} Object segmentation performance, measured in terms of foreground ARI~\cite{hubert1985comparing,greff2019multi} (FG-ARI$\uparrow$) in $\%$. We compare two recent state-of-the-art models, SAVi~\cite{anonymous2022conditional} (trained using optical flow) and SIMONe~\cite{kabra2021simone}. SAVi + BBox additionally receives object bounding boxes as cues in the first frame.}
\label{tab:object-discovery}

\begin{center}
\small{MOVi-A}\quad\,\,\,
\small{MOVi-B}\quad\,\,\,
\small{MOVi-C}\quad\,\,\,
\small{MOVi-D}\quad\,\,\,
\small{MOVi-E}\\[0.25em]
\begin{overpic} 
[width=\linewidth]
{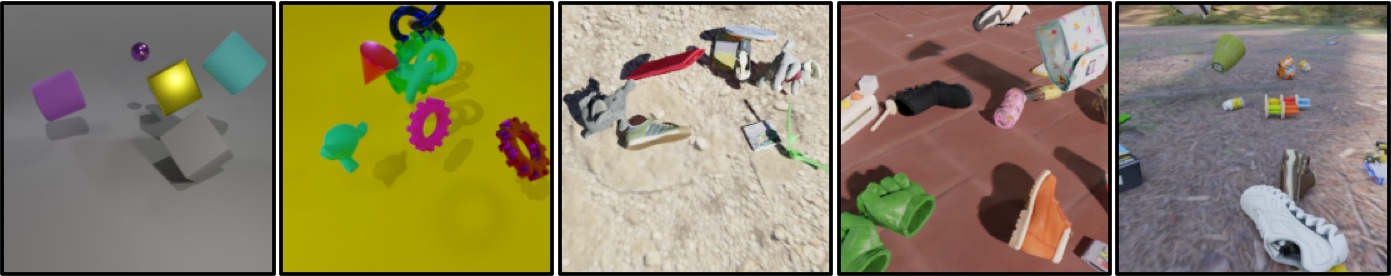}
\put(-4,4){\rotatebox{90}{\small{Frames}}}
\end{overpic}\vspace{0.25em}
\begin{overpic} 
[width=\linewidth]
{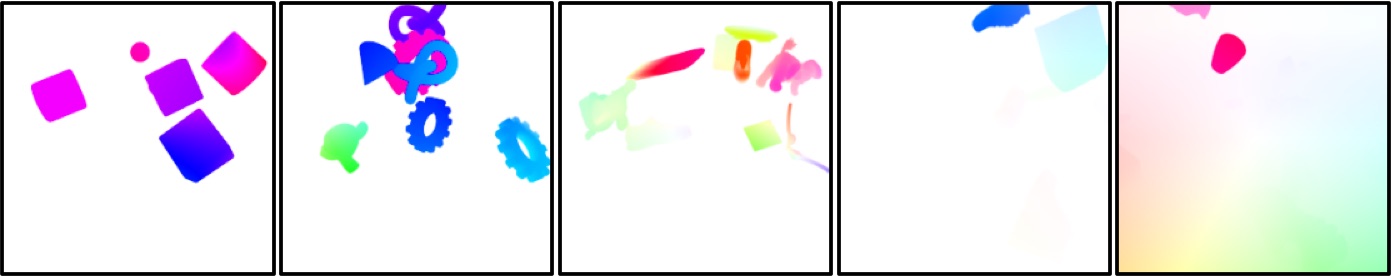}
\put(-4,7){\rotatebox{90}{\small{Flow}}}
\end{overpic}
\end{center}
\captionof{figure}{
\textbf{Object discovery --} 
Dataset samples of MOVi of increasing visual complexity. MOVi-A uses objects inspired by CLEVR~\cite{johnson2017clevr}. MOVi-B introduces additional primitive object types and colors. MOVi-C introduces real-world backgrounds and scanned 3D objects. MOVi-A to -C contain scenes of up to 10 moving objects (24 frames per video). MOVi-D \& MOVi-E scenes have up to 23 objects, with only a small fraction of moving objects. In MOVi-E, the camera is moving in random directions.}
\label{fig:object-discovery}
\vspace{-1em}
\end{figure}
\subsection{Object discovery from video}
\label{sec:object_discovery}
Object discovery methods aim to decompose a scene into its constituent components and find object instance segmentation masks with minimal supervision. While recent models such as IODINE~\cite{greff2019multi}, MONet~\cite{burgess2019monet}, GENESIS~\cite{engelcke2019genesis}, and Slot Attention~\cite{locatello2020object} succeed at decomposing simple scenes with uniform textures, decomposing dynamic scenes (i.e., videos) with high visual complexity and complex dynamics remains difficult.
This challenge introduces five Multi-Object Video (MOVi) datasets, MOVi-A to -E (see Fig.~\ref{fig:object-discovery}), of increasing visual and dynamical complexity, aimed at testing the limits of existing object discovery approaches, enabling progress towards more realistic and diverse visual scenes.

We test two recent state-of-the-art video object discovery methods, SAVi~\cite{anonymous2022conditional} and SIMONe~\cite{kabra2021simone}, for their ability to decompose videos into temporally consistent object segmentation masks (see Table~\ref{tab:object-discovery}). SAVi, which uses optical flow during training, performs better at decomposing videos with moving objects only, especially when receiving additional cues in the first frame of the video. Both methods decline in performance as complexity increases with an exception for static objects in MOVi-D and -E, which are sometimes partially captured by SIMONe. Neither method can reliably decompose scenes in all five datasets.  

\begin{figure}[t]
\centering
\resizebox{\linewidth}{!}{
    \begin{tabular}{lllccccc}
    &  &\phantom{a}& \multicolumn{2}{c}{Sintel } & \phantom{a} & \multicolumn{2}{c}{KITTI-15 } \\
    \cmidrule{4-5} \cmidrule{7-8} 
     Dataset  & Parameters && Clean & Final && AEPE & ER\%\\
    \midrule
    FlyingChairs (2D) & Manual && 2.27	&3.76 && 7.63  & 38.5\% \\
    Kubric (3D) & Manual && \bf{1.89} & {3.02} && {4.82} & {16.9}\%  \\ 
    AutoFlow (2D) & Learned && 2.08 & \bf{2.75} && \bf{4.66} & \bf{15.3}\%  \\      
    \bottomrule
    \end{tabular}}
\vspace{2.0mm}
\captionof{table}{
\textbf{Optical flow} --
Comparison of pre-training RAFT on different optical flow datasets (lower is better for all metrics).
}
\label{table:kubric_raft}
\begin{center}
\begin{overpic} 
[width=\linewidth]
{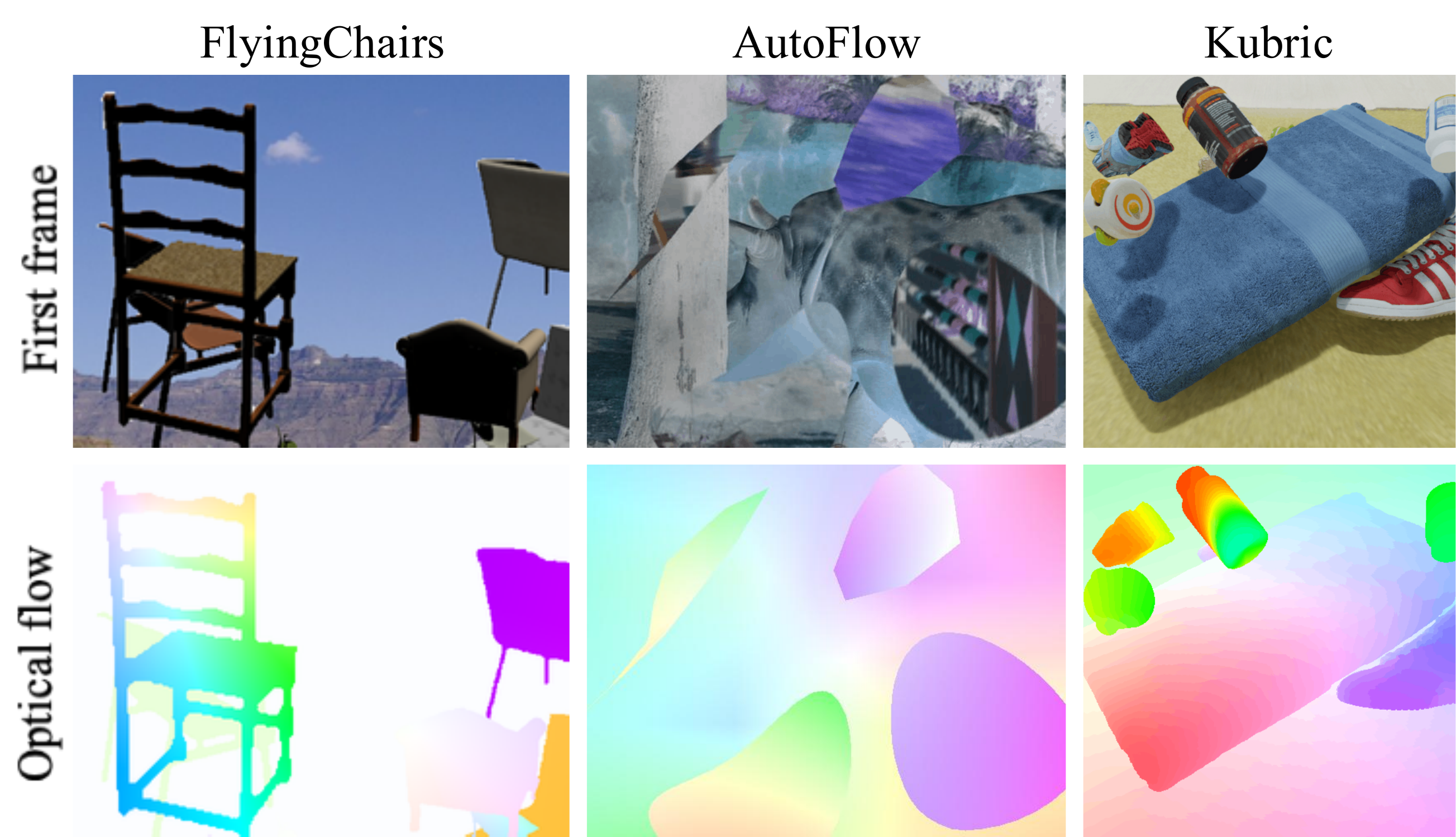}
\end{overpic}
\end{center}
\captionof{figure}{
\textbf{Optical flow} --
Chairs in FlyingChairs undergo 2D affine motion; random polygons  in AutoFlow  undergo nonrigid 2D motion; 3D objects in Kubric undergo 3D rigid-body motion.
}
\label{fig:opticalflow}
\vspace{-1em}
\end{figure}

\subsection{Optical flow}
\label{sec:optical_flow}
Optical flow refers to the 2D motion from pixels in one frame to the next in a video. It is fundamental to video processing and analysis. 
Unlike high-level vision tasks, we cannot obtain reliable, ground-truth optical flow on generic real-world videos, even with human annotation. 
Optical flow is actually the first sub-field of computer vision to rely on synthetic data for evaluation \cite{Barron94performanceof}.

Recent deep models,  PWC-net~\cite{sun2018pwc}, RAFT~\cite{teed2020raft}, and VCN~\cite{yang2019volumetric}, all rely on synthetic data for pre-training,
like FlyingChairs~\cite{FlowNet-ICCV2015}. 
However, FlyingChairs lacks photo-realism, uses synthetic chairs as the only foreground objects, and does not have general 3D motion.
AutoFlow \cite{sun2021autoflow} learns rendering hyperparameters for generating a synthetic flow datasets, yielding 
large performance gains over FlyingChairs~\cite{sun2021autoflow}. 
But  AutoFlow adopts a simple 2D layered model, lacks 3D motion and realism in rendering. 
Our Kubric dataset addresses these shortcomings, as shown in  Fig.~\ref{fig:opticalflow}. 

We compare training RAFT on different datasets using the same training protocol~\cite{teed2020raft,tf-raft,sun2021autoflow}.   
As shown in Table~\ref{table:kubric_raft}, Kubric leads to significantly more accurate results than FlyingChairs when both use manually selected rendering hyperparameters, demonstrating the benefit of using 3D rendering. 
Kubric also performs competitively against AutoFlow.
Note that this is not an apples-to-apples comparison, because the hyperparameters of AutoFlow have been learned to optimize the performance on the Sintel dataset~\cite{sun2021autoflow}. 
These results suggest that learning hyperparameters for Kubric is likely to result in significant performance gains.


\begin{figure}[t]
\begin{center}
\resizebox{0.9\linewidth}{!}{
\begin{tabular}{l|ccccc}
\toprule
Frequency Cutoff              & $10^{-0.5}$ & $10^{0}$ & $10^{0.5}$ & $10^{1}$ & $10^{2}$ \\
\midrule
PSNR $\uparrow$               & \bf{28.1}  & 27.8  & 26.7  & 23.6  & 23.4  \\
Depth Variance$\downarrow$    & 0.026 & 0.024 & 0.023 & 0.023 & \bf{0.022} \\
\bottomrule
\end{tabular}}
\vspace{1em}
\captionof{table}{{\bf Texture-structure in NeRF} -- Reconstruction error and depth variance for different texture frequency bands with NeRF on textured surface. Accuracy of color prediction improves as frequency of the texture becomes lower, while accuracy of the surface geometry degrades.}
\label{tab:nerf_texture}
\vspace{0.5em}
\end{center}
\begin{minipage}{\linewidth}
\begin{minipage}{.52\linewidth}
\includegraphics[width=\linewidth]{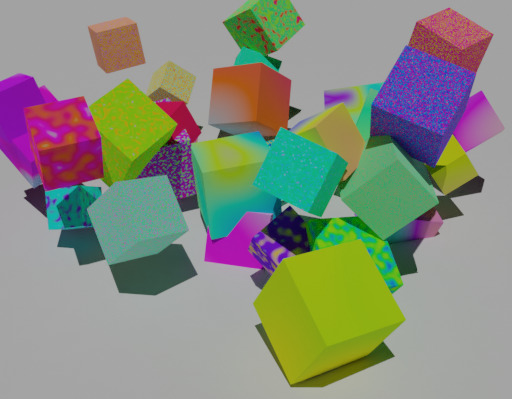}
\end{minipage}
\hfill
\begin{minipage}{.44\linewidth}
\captionof{figure}{
A NeRF dataset with procedural texture allows each pixel to be annotated with frequency information. This enables analysis of the frequency-structure relationship in the learned NeRF model.}
\label{fig:nerf_texture}
\end{minipage}
\end{minipage}
\vspace{-1em}
\end{figure}

\subsection{Texture-structure in NeRF}
\label{sec:texture_structure}
Neural radiance fields are inherently \emph{volumetric} representations, but are commonly used to model the surfaces of solid objects.
These NeRF surface models are a result of the model trying satisfy a multi-view reconstruction problem: to reconstruct surface detail consistently from multiple views, those details must lie in a thin slice of the volume around the true surface.
Note that not all surfaces will encourage NeRF to build a surface model.
Surfaces with flat color may still be reconstructed as a non-solid volume.
Hence, benchmarking NeRF methods according to how well they stay true to the actual surface depending on texture is an interesting aspect that is still unexplored.

To quantify this, we created synthetic scene containing flat surfaces, the textures of which are procedurally generated with blue noise to have varying spatial frequency.
We annotate each pixel with the cutoff frequency of its texture and analyze the correlation between frequency, depth variance, and reconstruction error.
We then train a NeRF model with this synthetic data.
As shown in \Table{nerf_texture}, we find that increasing frequencies are associated with lower depth variance, indicating better approximations to a hard surface, while also increasing the reconstruction error, showing that the network is less able to approximate the complex textures.
It would be interesting to see how well future volumetric multi-view reconstruction methods would cope with  this ambiguity and encourage hard surface boundaries.

\begin{figure}[t]
\begin{center}
\resizebox{\linewidth}{!}{
\begin{tabular}{@{}ccc@{}}
\toprule
Train data set & COCO + Active & COCO + Active + Synth \\ 
\midrule
COCO~\cite{lin2014microsoft}               &  0.554   &    0.557      \\
Active~\cite{movenet2021}             &  0.650    &    0.662      \\
Yoga               &  0.391    &    0.427       \\
\bottomrule
\end{tabular}
} 
\vspace{1em}
\captionof{table}{
\textbf{Pose estimation} --
results are improved out-of-domain by the addition of synthetic images of human models featuring poses outside the COCO domain; Keypoint Mean Average Precision~(mAP) metric~(higher is better)}
\label{tab:MoveNet}
\vspace{1em}
\begin{overpic} 
[width=\linewidth]
{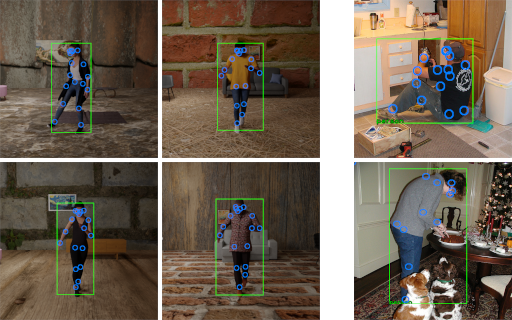}
\end{overpic}
\end{center}
\captionof{figure}{
\textbf{Pose estimation} --
fully annotated images from synthetic videos aimed at diversifying poses~(left), motions, subjects and backgrounds featured in real-world annotated data sets, and (right) examples of COCO-equivalent images.
}
\label{fig:simanim}
\vspace{-1em}
\end{figure}

\subsection{Pose estimation}
\label{sec:pose_estimation}
Pose-estimation-based interactive experience (e.g., Kinect) often feature human poses that remain under-represented in most data sets comprising user-generated pictures (e.g. COCO~\cite{lin2014microsoft}), as picture-worthy poses present an obvious sampling bias. Simulated data can supplement real data with less aesthetic poses which are nonetheless common in real-life human motions.
Here we improve MoveNet~\cite{movenet2021}, a CenterNet~\cite{zhou2019objects} based pose inference CNN usually trained on COCO~\cite{lin2014microsoft} and Active~\cite{movenet2021} (a proprietary data set with more diverse poses).
As in Simpose~\cite{zhu2020simpose}, training batches mix real and synthetic data with an $80/20\%$ mixture.
Unlike~\cite{zhu2020simpose}, synthetics do not provide additional labels (e.g., surface normals) but only contribute more diverse examples.
As illustrated in \Figure{simanim}, the samples feature $41$ rigged RenderPeople models placed in a randomized indoor scene where background elements and textures come from BlenderKit and TextureHeaven.
Human poses are extracted from dancing and workout ActorCore animations.
While licensing terms of non-CC0 assets forbid data publication, the data set can be re-generated with our open source software by any owner of the same mesh and animation assets.
Synthetic data improves keypoint Mean-Average-Precision (see Table~\ref{tab:MoveNet}), in domain (on COCO and Active), and out-of-domain (on Yoga, a test set of contorted poses comprising 1000 examples).
Synthetic data are therefore now routinely used in our human-centric training procedures for still images as well as videos.


\begin{figure}[t]
    \centering
    \hspace{6pt}\includegraphics[width=0.93\linewidth]{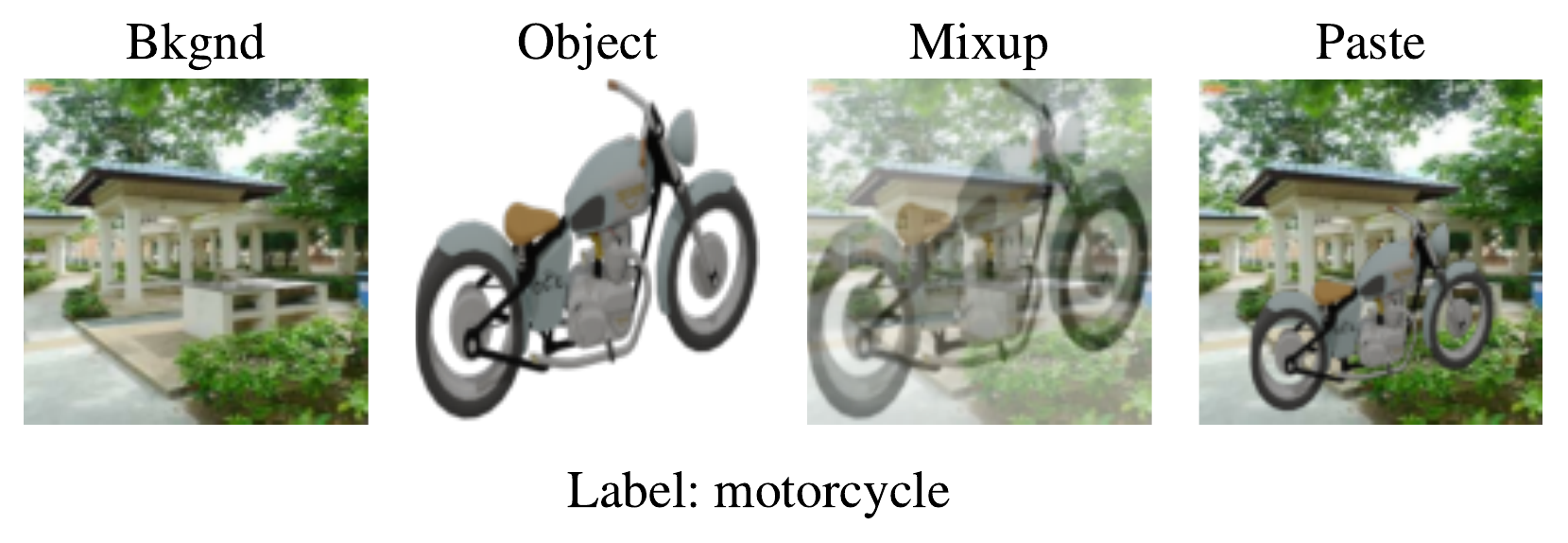}
    \includegraphics[width=0.99\linewidth]{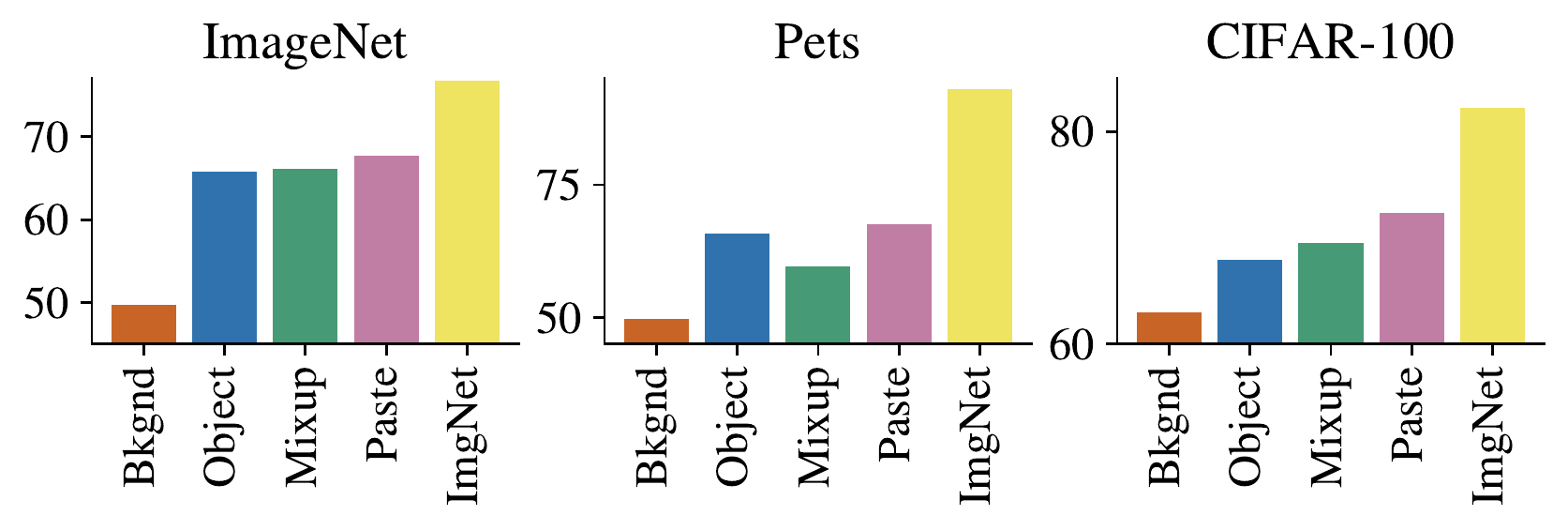}
    \vspace{10pt}
    \caption{\textbf{Pre-training} a ResNet50 on synthetic Kubric data (top) and transferring it to standard bencharks (bottom) halves the gap between random pre-training (Bkgnd) and ImageNet pre-training.}\label{fig:pretrain}
    \vspace{-1em}
\end{figure}

\subsection{Pre-training visual representations}
\label{sec:pretraining}
Ever since AlexNet~\cite{alexnet}, the entire field of computer vision has benefited immensely from re-using ``backbones'' pre-trained on large amounts of data~\cite{cnn_off_shelf,bit,vit,simon_transfer}.
However, recent work casts doubt on the  continued use of datasets that consist of vast collections of photos from the internet~\cite{i21k_fairness,pyrrhic}.
One potential way forward, which completely circumvents the downsides of web-image based data, is to use  rendered data.
This has recently shown great success for face recognition~\cite{wood2021fake}, and we hypothesize that synthetic data could also eventually replace web images for pre-training general computer vision backbones.
In order to evaluate the promise of such a setting, we perform a small pilot experiment. Kubric was used to 
render ShapeNet objects in various random poses on transparent backgrounds.
We pre-train a ResNet-50 to predict the object's category from images that combine the object with a random background image in various ways, as shown in Fig.~\ref{fig:pretrain}~(top).
We then transfer this pre-trained model to various datasets, following the protocol in~\cite{bit}.
Figure~\ref{fig:pretrain}~(bottom) shows that this simple pilot experiment already halves the gap between random pre-training and pre-training on ImageNet, suggesting that this is a promising approach.


\begin{figure}[t]
\centering
\resizebox{.7\linewidth}{!}{ 
\begin{tabular}{@{}lc|cc@{}}
& static & teleport & jitter \\
\midrule
NeRF-L2 &        43.5~dB & 25.5~dB & 27.4~dB\\
NeRF-L1 &     42.8~dB & 40.4~dB & 37.1~dB\\
\end{tabular}
} 
\vspace{1em}
\captionof{table}{
\textbf{\RobustNeRF --}
Performance in PSNR$\uparrow$ of classical NeRF-L2~\cite{nerf}) and its L1-robust version~\cite{vonis}.
Note the performance is evaluated on test views (i.e.~novel view synthesis) and \textit{without} any impostors present.
}
\label{tab:robustnerf}
\vspace{.5em}
\begin{center}
\begin{overpic} 
[width=.9\linewidth]
{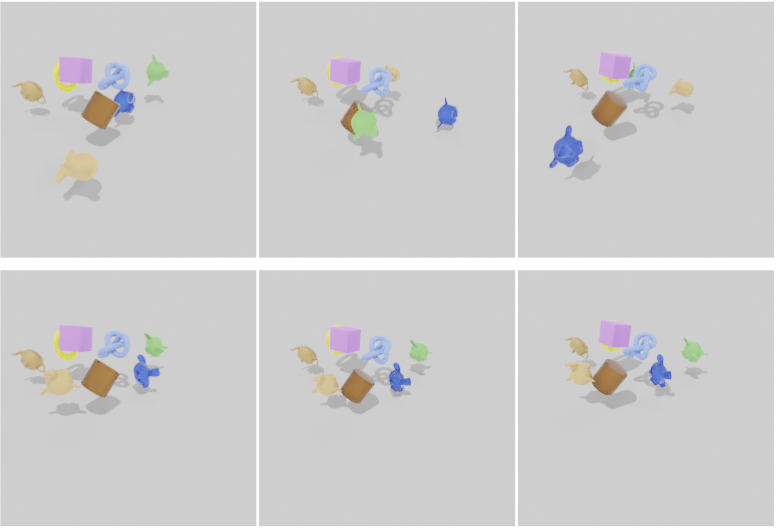}

\put(8,69){Frame 241}
\put(40,69){Frame 251}
\put(73,69){Frame 261}
\put(-5,44){\rotatebox{90}{Teleport}}
\put(-5,14){\rotatebox{90}{Jitter}}
\end{overpic}
\end{center}
\captionof{figure}{
\textbf{\RobustNeRF (training data)} --
a dataset that violates the rigidity assumptions typically assumed by NeRF training workloads.
Here, we qualitatively visualize the ``teleport'' and ``jitter'' versions of the training dataset.
}
\label{fig:robustnerftrain}
\begin{center}
\vspace{.1in}
\begin{overpic} 
[width=.9\linewidth]
{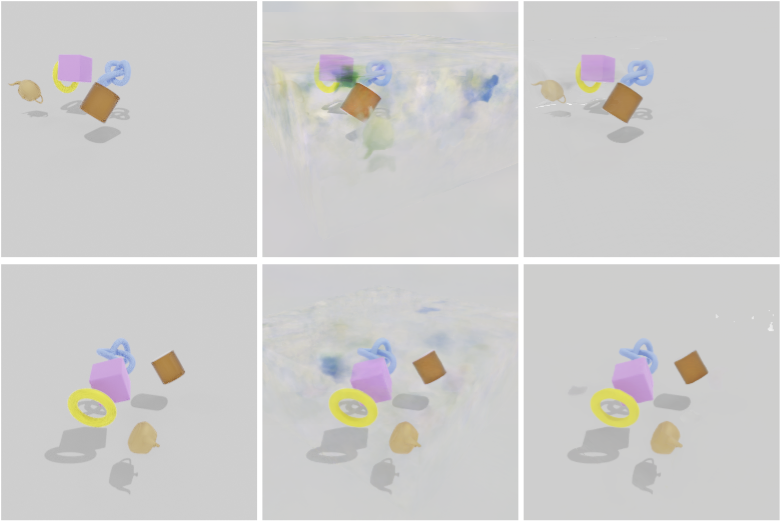}

\put(4,68){Ground Truth}
\put(42,68){NeRF-L2}
\put(74,68){NeRF-L1}
\put(-5,44){\rotatebox{90}{Teleport}}
\put(-5,14){\rotatebox{90}{Jitter}}
\end{overpic}
\end{center}
\captionof{figure}{
\textbf{\RobustNeRF (training outcome)} --
Visualizing the rendering of NerF-L2 vs Nerf-L1 for models trained on Teleport and Jitter training sets. During test the ground truth does not have dynamic objects. Typical NeRF-L2 models render a shadow in place of transient objects where as the NeRF-L1 model can successfully remove the floaters.
}
\label{fig:robustnerftest}
\end{figure}
\subsection{\RobustNeRF} 
\label{sec:robust_nerf}
Neural Radiance Fields~\cite{nerf} or NeRF, trains a representation of a static 3D scene via volume rendering by minimizing a photometric reconstruction loss of the form $\loss{}(\theta) = \expect_r \| I(r) - I_\theta(r) \|_2^2$, where $r$ are rays corresponding to pixels of a multi-camera system.
The nature of this loss implies that when the scene is \textit{not perfectly static} across views, the recovered representation is corrupted; see~\Figure{robustnerftest}~(center).
This challenge demonstrates that further research is still needed to fully address this problem; see~\Table{robustnerf}.
In the~``teleport'' challenge, while most of the scene remains rigid, we add impostor non-static object (i.e. the monkey head) randomly within the scene bounds, while in the~``jitter'' challenge the impostor position jitters around a fixed position.
In other words, the two datasets evaluate the sensitivity of unstructured (teleport) vs. structured (jitter) outliers in the training process. \Figure{robustnerftrain} showcases some of the training frames for each challenge.
As shown in \Table{robustnerf}, while unstructured outliers are \textit{to some degree} addressed by NeRF-L1~($-2.4$~dB), structured outliers are significantly more challenging to overcome. 
\newcommand{\SOD}{{Salient Object Detection}\xspace}
\subsection{Multi-view object matting}
\label{sec:multi_view_sod}

\begin{figure}[t]
\begin{overpic}[width=\columnwidth]
{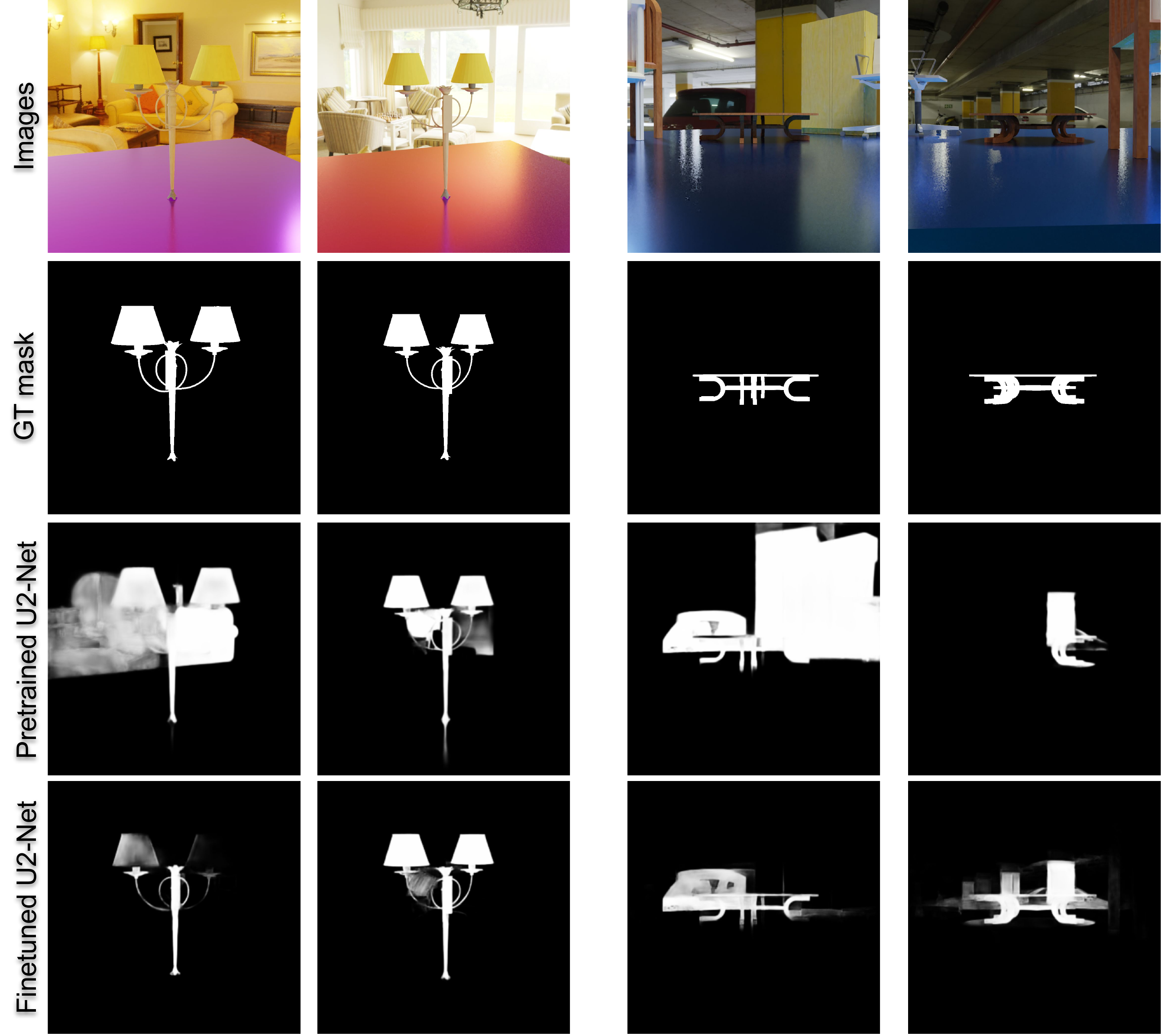}
\put(17,-4){SOD ``easy''}
\put(66,-4){SOD ``hard''}
\end{overpic}
\vspace{1.0em}
\caption{
\textbf{\SOD} -- 
Images, ground truth mask, as well as predictions from $U^2$-Net (pretrained and fine-tuned) over an example scene in the easy and hard datasets.}
\label{fig:sod}
\end{figure}

\begin{table}
\small
\addtolength{\tabcolsep}{-2.5pt}
\begin{tabular}{lccccccc}
& \multicolumn{3}{c}{easy} && \multicolumn{3}{c}{hard} \\
\cmidrule(lr){2-4} \cmidrule(lr){6-8}
&  maxF$_\beta$              & MAE            & S$_m$ & \hspace{0.1cm}&  maxF$_\beta$              & MAE            & S$_m$\\
\hline
\multicolumn{2}{l}{\textbf{Pre-trained}}                &            &             \\
\hspace{0.1cm} SINet \cite{camouflaged_fan2020}    & 0.494             & 0.097             & 0.597         && 0.401         & 0.093         & 0.568           \\
\hspace{0.1cm} EGNet \cite{egnet_zhao2019}         & 0.652             & 0.090             & 0.753         && 0.462         & 0.133         & 0.641                \\
\hspace{0.1cm} BASNet \cite{basnet_qin2019}        & 0.822             & 0.034             & 0.878         && 0.581         & 0.086         & 0.730     \\
\hspace{0.1cm} CPD \cite{cpd_wu2019}               & 0.811             & 0.029             & 0.872         && 0.594         & 0.068         & 0.742    \\
\hspace{0.1cm} $U^2$-Net \cite{u2net_qin2020}      & 0.825             & 0.032             & 0.882         && 0.594         & 0.083         & 0.743     \\
\hline
\multicolumn{2}{l}{\textbf{Fine-tuned}}                    &                  &                    \\
\hspace{0.1cm} CPD \cite{cpd_wu2019}               & 0.958             & 0.007             & 0.968         && 0.901         & 0.015         & 0.925     \\
\hspace{0.1cm} $U^2$-Net \cite{u2net_qin2020}      & 0.975             & 0.006             & 0.977         && 0.909	        & 0.015	        & 0.931     \\		
\hline
\multicolumn{2}{l}{\textbf{Trained from scratch}}                   &                  &                    \\
\hspace{0.1cm} CPD  \cite{cpd_wu2019}              & 0.957             & 0.008             & 0.967         && 0.906         & 0.015         & 0.928\\
\hspace{0.1cm} $U^2$-Net \cite{u2net_qin2020}      & 0.967             & 0.009             & 0.971      && 0.890	        & 0.021	        &0.916
\end{tabular}
\vspace{1em}
\caption{\textbf{\SOD} -- Quantitative results as evaluated by max F-measure with $\beta^2=0.3$~\cite{achanta2009frequency}, Mean Absolute Error~\cite{mae_borji2015} and Structure measure~\cite{smeasure_fan2017}.}
\label{tab:sod_easy}
\end{table}


\SOD~(SOD) aims to segment out the most salient object in an image from the background.
Classical methods involve active-contour~\cite{osher1988fronts, kass1988snakes} or graph-cut~\cite{wu1993optimal, cox1996ratio} techniques, but there also exist techniques with human-in-the-loop~\cite{russell2008labelme, wang2014touchcut, qin2018bylabel}, and more recently deep learning variants~\cite{u2net_qin2020, basnet_qin2019, egnet_zhao2019, camouflaged_fan2020, cpd_wu2019}.
With human feedback, interactive methods are typically robust, but also costly.
Automatic segmentation, be it with traditional methods or deep networks, are less performant.
Here we propose a new mode of operation in SOD, a significantly harder task (see~Figure~\ref{fig:sod}), yet with sufficient information for a human to solve the problem without ambiguity.
We compare several single view state-of-the-art SOD algorithms on this dataset, and propose two datasets with increased complexity.
Instead of one image, we assume access to \textit{multiple} images~(taken from different angles) of the same salient object.
With multiple views of the same object, we theorize that automatic SOD would be more robust as the 3D structure implied from multiple images provides information that could help disambiguate boundaries of the target object.
For the \textit{easy} challenge, scenes only contain \textit{one} salient object within the scene, while in the \textit{hard} challenge we additionally insert clutter.
All target objects are randomly selected from \textsc{ShapeNetCore~V2} the background from~\textsc{Polyhaven}~HDRIs~\cite{polyhaven2021}.
In the case of the \textit{hard} challenge, clutter objects are also sampled randomly from \textsc{ShapeNetCore~V2}.
We render~$10$~images for each scene, and export images and segmentation masks with Kubric.
The training/test sets contains~$1000/100$~scenes respectively for both \textit{easy} and \textit{hard}.

To the best of our knowledge, multi-view SOD baselines do not exist. We therefore evaluate recent SOTA single-view SOD models as baselines.
We first evaluate the pretrained models on the \textit{easy} dataset, as summarized in Table~\ref{tab:sod_easy}.
Some pretrained models (e.g. $U^2$-Net) performs decently.
Next, we fine-tune the best performing pretrained models ($U^2$-Net and CPD) on the \textit{easy} dataset.
Despite $U^2$-Net and CPD's strong performance, however, multivew enabled SOD models should be stronger as the failure cases of single view SOD models (e.g. see Figure~\ref{fig:sod}) indicate the lack of 3D understanding is often the culprit.
Last but not least, we train $U^2$-Net and CPD on the \textit{easy} dataset from scratch for additional baseline results; see Table~\ref{tab:sod_easy}.
\quad
We repeat the experiments on the \textit{hard} dataset. 
In the presence of clutter, the task becomes significantly harder.
Clutter is often mistaken to be the salient object especially when the objects are in close proximity.
Again, the lack of 3D understanding is an important factor for the relative poor performance of the models.
Models that work with multiple views, we hypothesize, will significantly improve upon the baselines.

\begin{figure}[t]
\begin{center}

\resizebox{.8\linewidth}{!}{ 
\begin{tabular}{lcccc}
 & \multicolumn{2}{c}{Lambertian} & \multicolumn{2}{c}{Specular} \\
\cmidrule(lr){2-3} 
\cmidrule(lr){4-5}
& PSNR & SSIM  & PSNR & SSIM\\
\midrule
LFN~\cite{sitzmann2021lfns}  & 29.71 & 0.883 & 26.77   & 0.816 \\
PixelNeRF~\cite{yu2021pixelnerf} & \textbf{32.11}  & \textbf{0.922} & \textbf{29.96} &  \textbf{0.885}\\      
\end{tabular}
} 
\vspace{1em}
\captionof{table}{
\textbf{Complex BRDFs --}
Comparison of novel view synthesis results for specular vs. diffuse materials.
}
\label{tab:brdf}
\vspace{1em}
\begin{overpic} 
[width=\linewidth,height=0.37\linewidth]
{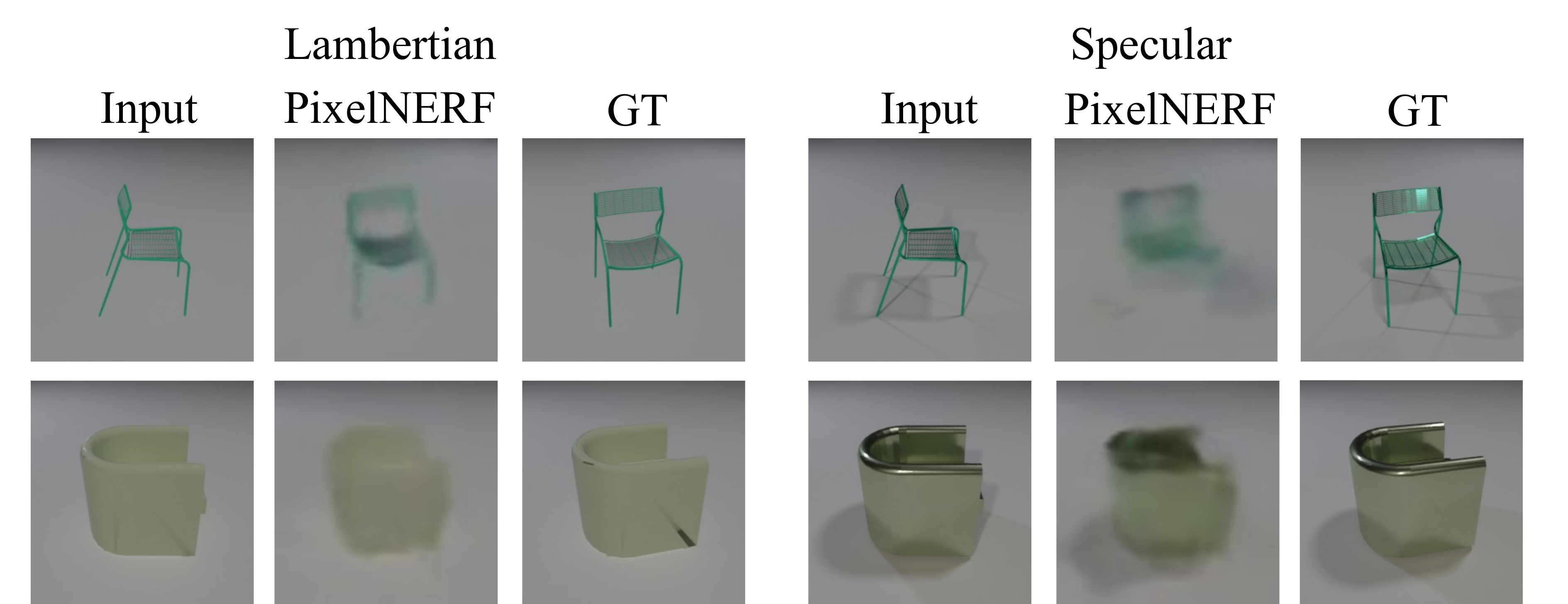}
\end{overpic}
\end{center}
\captionof{figure}{
\textbf{Complex BRDFs --}
Existing approaches struggle to model ShapeNet shapes rendered with specular materials.
}
\label{fig:brdf}
\end{figure}

\subsection{Complex BRDFs}
\label{sec:complex_brdfs}
Consider the core vision problem of reconstructing a 3D scene from few observations~\cite{sitzmann2019srns, yu2021pixelnerf, niemeyer2020dvr, sitzmann2021lfns}.
Current datasets ~\cite{kato2018neural,yu2021pixelnerf} mostly feature Lambertian scenes, i.e., scenes that consist of mostly diffuse surfaces, with few specular highlights.
In this case, the only relevant scene parameters are the 3D geometry, as well as the \textit{diffuse} surface color. 
When scene surfaces are highly reflective, the number of scene properties required for accurate novel view synthesis grows significantly.
Instead of just 3D shape and appearance, the model needs to address 3D geometry, the BRDF of every surface point, as well as a full characterization of the light incident onto the scene. 
To this end, we render out a highly specular version of the ShapeNet dataset as a challenge for few-shot novel view synthesis algorithms.
We follow Kato~et~al.~\cite{kato2018neural} and render objects across 13 classes from the  same 24 views.
To each object, we randomly assign an RGB color. We place three light sources at randomized positions on the upper hemisphere.
In this challenge, we fix the material properties of each object to the properties of the specular CLEVR~\cite{johnson2017clevr}, and ray-trace each scene with~$12$~ray bounces.
We benchmark two recently proposed models on this dataset: Light Field Networks~\cite{sitzmann2021lfns}, which parameterizes a scene via its 360-degree light field, and PixelNeRF~\cite{yu2021pixelnerf}, a conditional 3D-structured neural scene representation.
In order for these models to successfully train and perform at test-time, they need to both model the view-dependent forward model correctly, and correctly infer the position of the light sources.
We find that both models perform substantially worse in Table~\ref{tab:brdf} and specular compared to Lambertian shapes.
In Figure~\ref{fig:brdf}, we illustrate how existing approaches struggle to represent inherent specularities in shapes.

\newcommand{\SVR}{Single View Reconstruction\xspace}

\begin{figure*}[t]
\begin{center}
\includegraphics[width=\linewidth]{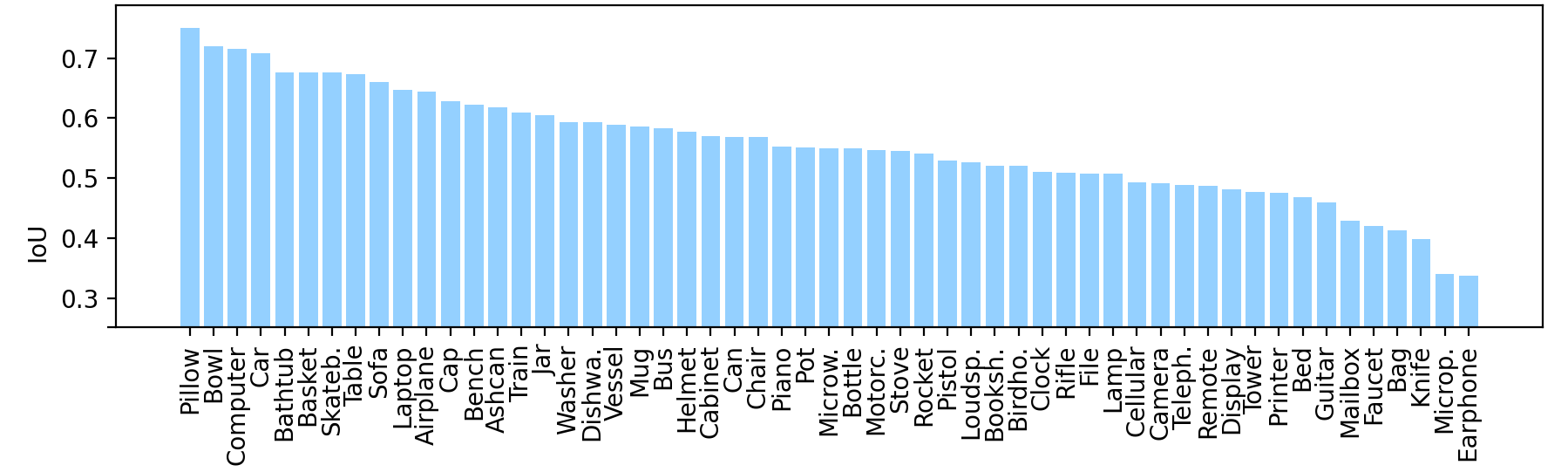}
\vspace{0.05em}
\captionof{figure}{
\textbf{\SVR} --
IoU results of training SoftRas on all ShapeNet categories. Pillows and bowls have the highest IoU and the model struggle's most with microphones and earphones.
}
\label{fig:shapenet_quantitative}
\end{center}
\end{figure*}

\vspace{1em}
\begin{figure}[t]
\begin{center}
\includegraphics[width=\linewidth]{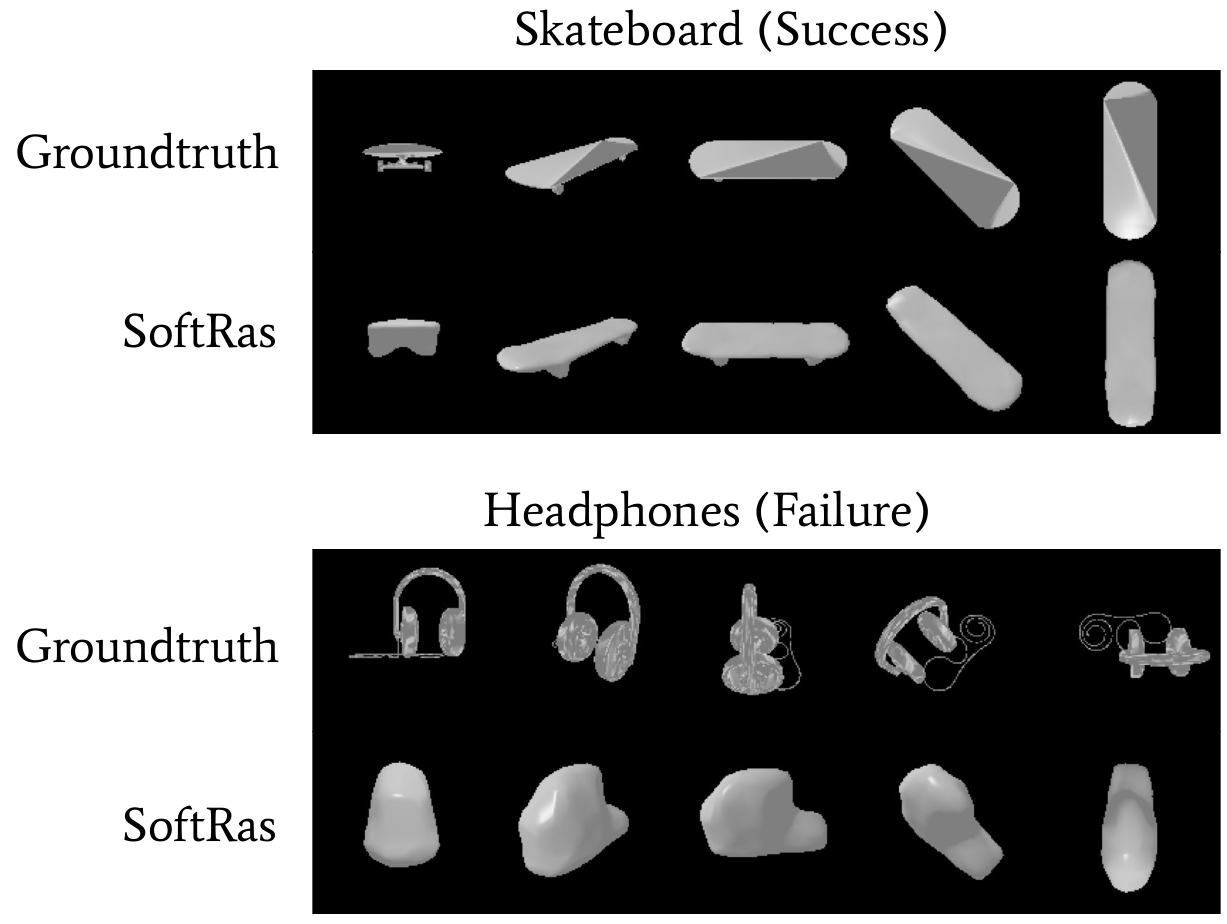}
\vspace{0.25em}
\captionof{figure}{
\textbf{\SVR} -- 
Qualitative results of SoftRas trained on the full ShapeNetCore~V2 dataset of around 51,300 objects. }
\label{fig:shapenet_qualitative}
\end{center}
\end{figure}

\subsection{\SVR}
\label{sec:recons3d}
%
Reconstructing an explicit 3D representation of an object~(e.g.~polygonal mesh) exclusively from 2D image supervision is challenging due to the ill-posed nature of the problem. Given input 2D images and their associated 3D viewpoint parameters (i.e., comprising the azimuth, distance, and elevation of the camera looking at the object), current methods (e.g.~\textsc{SoftRas}~\cite{liu2019soft}) combine an encoder to first extract latent features from images followed by a decoder to extract 3D vertices and face connectivity from the encoded feature vectors. Next, a differentiable renderer can project the 3D faces according to the viewpoint and all while respecting view-dependent occlusion constraints. To train such a rasterization-based differentiable rendering model, a loss function can be formulated from the difference between this projected (i.e., rendered) output and an image of the silhouette of the object against ground-truth viewpoints. One common loss function is based on a soft IoU loss between the projected and ground-truth images. Notably, this entire optimization no longer relies on any direct 3D supervision on the explicit 3D object parameterization, only viewpoint labels are needed and these can be readily determined from sourcing camera responsible for producing the 2D supervision images.

We train a \textsc{SoftRas}~\cite{liu2019soft} model on the \textit{entire} \textsc{ShapeNetCoreV2} dataset instead of the commonly-used subset of \textsc{ShapeNet} that only has 13-category, that's typical for published work in this area \cite{liu2019soft,kato2018neural,yan2016perspective}. The full \textsc{ShapeNetCoreV2} consists of 55 categories with a total of approx. $51, 300$ object models.
We leverage Kubric's ability to automatically process these object models and project each into~$24$ random viewpoints, all while maintaining consistent meta information (camera pose and object category) that allows us to train \textsc{SoftRas} efficiently. 
We trained SoftRas on two experimental setups: ``in-distribution'', for which we follow the training regimen of \cite{liu2019soft}, train on 80\% of each category, and test and report performance on the remaining 20\% of each category, and ``out-of-distribution'' where we train on all categories except 4 classes that we leave out for testing. They are {\it train, tower, washer and vessel}. 
Our results for ``in-distribution'', summarized in Figures~\ref{fig:shapenet_quantitative} and~\ref{fig:shapenet_qualitative}, illustrate that we perform best on pillows and bowls (IoU 0.75 and 0.72), and worst on microphones and earphones (IoU 0.34). For ``out-of-distribution'' the results on the test classes are close to those reported in Figure~\ref{fig:shapenet_quantitative}, suggesting that SoftRas can generalize to new classes, but its limitations are on reconstructing images from classes with complex shapes like {\it headphones}. We observe that this processed dataset allows us to train a \textsc{SoftRas}   capable of reconstructing a \textit{wider range of objects} than in the original work of Liu et al.~\cite{liu2019soft} but the performance for some classes are poor, which will hopefully inspire further research into more powerful and 2D-to-3D reconstruction methods. 
\newcommand{\VBR}{Video Based Reconstruction\xspace}

\begin{figure}[t]
\begin{overpic} 
[width=\linewidth]
{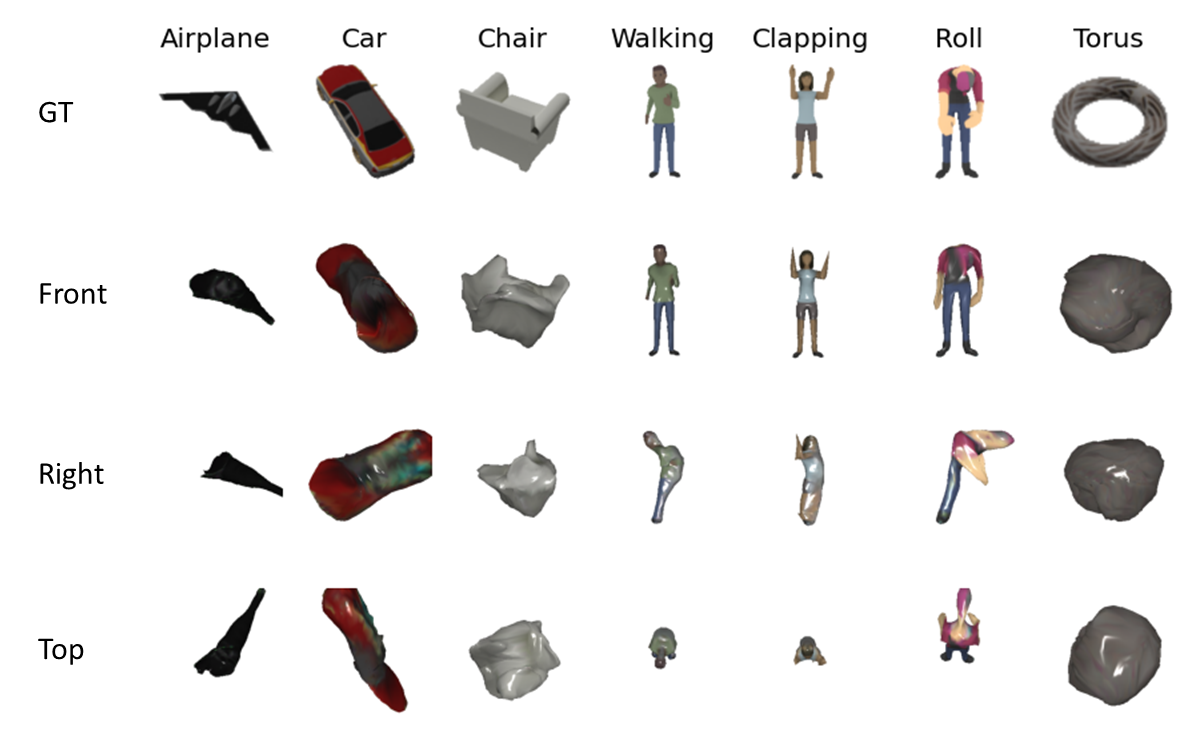}
\end{overpic}
\vspace{-1.0em}
\captionof{figure}{
\textbf{\VBR} --
Example results of 3D mesh reconstruction using LASR~\cite{yang2021lasr}.
For each object class, we challenge LASR with a Kubric-generated 30-frame video consisting of the color image, object silhouette, and optical flow for each frame.
We show the input (training) view of the object in the first row (GT). Others show the test views of the reconstructed object rendered from multiple viewing positions. Note that only the second row (Front) shares the same camera view as the input.
}
\label{fig:lasr}
\vspace{1em}
\centering
\begin{adjustbox}{width=\columnwidth,center}
\begin{tabular}{@{}ccccccc@{}}
\toprule
Input            & airplane &  car & chair & walking & clapping & roll \\ \midrule
True flow       & \bf{0.62} & \bf{0.22} & \bf{1.00} & 3.81 & \bf{1.30} & \bf{1.75} \\
Estimated flow  & 1.52 & 0.26 & 1.08 & \bf{3.19} & 1.34 & 1.85\\ 
\bottomrule
\end{tabular}
\end{adjustbox}
\vspace{0.5em}
\captionof{table}{
\textbf{\VBR} --
Mesh reconstruction error measured by Chamfer Distance for each object with either the ground truth optical flow provided by Kubric or the estimated flow as training input.
}
\label{tab:LASR}
\end{figure}

\subsection{\VBR}
\label{sec:robust3d_recons}
As discussed in \Section{recons3d}, the ill-posed nature of single-shot 3D reconstruction makes it an extremely challenging task.
To better supervise surface reconstruction, the extensive availability of video data provides an attractive alternative to images.
Multi-frame consistency of video sequences imposes additional constraints.
However, since most 3D scenes are not static and many interesting real-world objects are not rigid, it brings up a new challenge for video-based surface reconstruction methods to be robust to deformations.

LASR~\cite{yang2021lasr} presents a pipeline that jointly recovers object surface mesh, articulation, and camera parameters from a monocular video without using category-specific shape templates.
The method first uses off-the-shelf networks to generate a mask (silhouette) of the main object and optical flow for each frame.
Then, by leveraging~\textsc{SoftRas}~\cite{liu2019soft}, LASR jointly optimizes the object's rest shape, articulation, skinning weights, and camera parameters by minimizing the difference between the input and re-rendered color image, silhouette, and optical flow for each frame.

In this challenge, we first leverage Kubric to generate videos of both rigid (ShapeNet assets) and non-rigid \footnote{Human rigs imported from \url{https://quaternius.com}.} objects to evaluate the general performance of LASR. As shown in~Figure~\ref{fig:lasr}, LASR fits the mesh well with input views but fails to extrapolate to unseen views.
As the optical flow loss has been demonstrated to be critical and the ground truth flow can never be obtained from real data, we also evaluate how much LASR relies on the accuracy of the flow estimation.
We separately train LASR using either the estimated optical flow or the ground truth provided by Kubric and compare the results of reconstruction; see~Table~\ref{tab:LASR}.
As expected, training with the ground truth optical flow improves performance, although this improvement is marginal.
Another fundamental limitation of LASR, as with many other mesh-based differentiable renderers, is that it assumes a \textit{fixed} mesh topology and thus cannot handle topological changes. In Figure~\ref{fig:lasr}, we show an example where LASR fails to reconstruct a non zero genus shape~(i.e.~torus), highlighting the need for additional research towards more robust approaches. 
\newcommand{\PointTracking}{Point Tracking\xspace}
\begin{figure*}[t]
\resizebox{\linewidth}{!}{ 
\begin{tabular}{l|ccc|ccccc|ccccc}
\toprule
method &  AJ & $<\delta^{x}_{avg}$ & OA & Jac. $\delta^{0}$ & Jac. $\delta^{1}$ & Jac. $\delta^{2}$ & Jac. $\delta^{3}$ & Jac. $\delta^{4}$ & $<\delta^{0}$ & $<\delta^{1}$ & $<\delta^{2}$ & $<\delta^{3}$ & $<\delta^{4}$\\
\midrule
Na{\"i}ve & 28.6 & 44.7 & 82.1 &  9.7 & 16.2 & 25.8 & 38.5 & 52.9 & 18.2 & 28.6 & 42.6 & 59.0 & 74.8 \\
Contrastive & 49.5 & 68.7 & 80.5 & 17.2 & 35.5 & 53.9 & 67.1 & 73.6 & 30.8 & 55.1 & 75.0 & 88.0 & 94.2 \\ 
\bottomrule
\end{tabular}
} 
\vspace{1em}
\captionof{table}{
\textbf{\PointTracking} --
Performance of our contrastive point tracking baseline.
Jac.~$\delta^{x}$ is the Jaccard metric measuring both occlusion estimation and point accuracy, with a threshold of $\delta^{x}$;
AJ is the Average Jaccard across $x$ between $0$ and $4$.  
$<\delta^{x}$ is the fraction of points not occluded in the ground truth for which the prediction is less than $\delta^{x}$, and~$<\delta^{x}_{avg}$ is the average across $x$ between $0$ and $4$.
Occlusion Accuracy is denoted with OA. 
We set $\delta=2$.
}
\label{tab:point_tracking}
\end{figure*}

\begin{figure*}[t]
\resizebox{\linewidth}{!}{ 
\begin{tabular}{l|ccc|ccccc|ccccc}
\toprule
method &  AJ & $<\delta^{x}_{avg}$ & OA & Jac. $\delta^{0}$ & Jac. $\delta^{1}$ & Jac. $\delta^{2}$ & Jac. $\delta^{3}$ & Jac. $\delta^{4}$ & $<\delta^{0}$ & $<\delta^{1}$ & $<\delta^{2}$ & $<\delta^{3}$ & $<\delta^{4}$\\
\midrule
Na{\"i}ve & 28.6 & 44.7 & 82.1 &  9.7 & 16.2 & 25.8 & 38.5 & 52.9 & 18.2 & 28.6 & 42.6 & 59.0 & 74.8 \\
Contrastive & 45.2 & 63.6 & 80.2 & 12.6 & 28.9 & 48.8 & 64.1 & 71.8 & 23.7 & 47.0 & 69.9 & 85.1 & 92.4 \\ 
\bottomrule
\end{tabular}
} 
\vspace{1em}
\captionof{table}{
\textbf{\PointTracking} --
Performance for on vertically flipped videos from the evaluation dataset.  We see a roughly $4\%$ loss in performance on the average Jaccard score, suggesting that our method somewhat overfits to the scene layout.  However, the performance is far from collapsing.
}
\label{tab:point_tracking_vflip}

\end{figure*}

\begin{figure*}[t]
\begin{center}
\begin{tabular}{@{}c@{}c@{}c@{}c@{}}
\includegraphics[width=.22\textwidth]{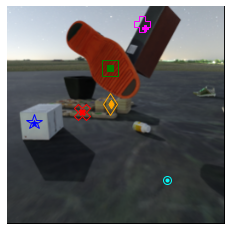}&\includegraphics[width=.22\textwidth]{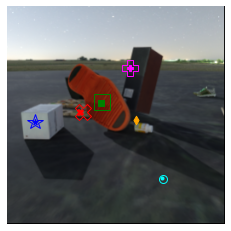}&\includegraphics[width=.22\textwidth]{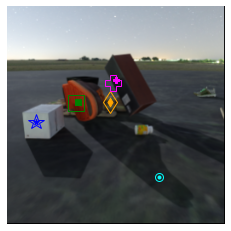}&\includegraphics[width=.22\textwidth]{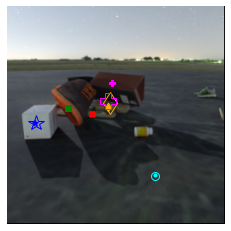}\\[-.5em]
frame 3 & frame 7 & frame 11 & frame 15 \\       
\end{tabular}
\end{center}
\vspace{0em}
\captionof{figure}{
\textbf{\PointTracking} --
Visualization of points tracked using our contrastive learning algorithm.  We show each of 6 tracked points with a different symbol and color, where the smaller, filled symbol is the prediction, while the larger, unfilled symbol is the ground truth.  If the ground truth is occluded or the point is predicted to be occluded, the corresponding symbol is not shown.  We show only frames 3, 7, 11, and 15 out of a full 24-frame sequence for brevity.  Note that the query frame is not necessarily at the beginning.  The query was in frame 3 for the cyan circle, the green square, and the red X; frame 7 for the blue star and the magenta plus, and frame 11 for the orange diamond.
}

\label{fig:point_tracking}
\end{figure*}

\subsection{\PointTracking}
\label{sec:point_tracking}
%
The concept of optical flow can be easily extended to longer-term tracking, following the approach proposed in Tracking Any Point~\cite{tapnet}.
That is, given a video and a target point on a surface in the scene~(specified in 2D), the goal is to output the 2D locations where that point appears in other video frames.
This kind of motion inference has many applications, from 3D surface reconstruction, to inference of physical properties like center of gravity and elasticity, to memory of objects across long episodes for an interactive agent.
Optical flow is insufficient in many of these domains because it cannot deal with occlusion, and small errors on frame pairs can lead to drift, resulting in large errors over time.  
It is straightforward to obtain long-term tracks from Kubric by identifying a point on a 3D object, and then projecting that point throughout the scene. Large-scale training datasets for this problem are very difficult for humans to annotate, so synthetic data can serve a critical role in achieving good performance.

Given a video, the annotations consist of a set of trajectories, i.e. a set of 2D points~$(x_{i,t},y_{i,t}) \in \mathcal{R}^{2}$ where $i$ indexes the different tracked points, and $t$ indexes time.
The ground truth also includes~$v_{i,t} \in \{0,1\}$, which indicates whether a point $i$ is visible in frame $t$ is visible ($v_{i,t}=1$) or not ($v_{i,t}=0$).
The tracking algorithm receives one visible point~$(x^{*}_{i},y^{*}_{i},t^{*}_{i})$ for each trajectory, and must output an estimate~$(\hat{x}_{i,t},\hat{y}_{i,t})$ for the other frames, as well as a visibility prediction~$\hat{v}_{i,t}\in \{0,1\}$.

\paragraph{Metrics}
%
We use three metrics proposed for Tracking Any Point~\cite{tapnet}.  The first ignores the output positions and evaluates occlusion estimation alone, via a simple classification accuracy which gives equal weight to each point on each frame.  The second metric evaluates only tracking accuracy.  Frames marked as occluded in the ground truth are ignored.  For the rest, we report a PCK-style~\cite{andriluka20142d} accuracy across several different thresholds.  That is, for a given threshold $\alpha$, which is a distance in pixels, we consider a point correct~if~$\sqrt{(x_{i,t}-\hat{x}_{i,t})^2+(\hat{y}_{i,t}-y_{i,t})^2}<\alpha$.
Our final metric combines both classification accuracy and detection accuracy, and is inspired by the Jaccard-style metrics from the object tracking literature~\cite{luiten2021hota}.
Let $TP_{\alpha}$ be the set of true positives: that is, all visible ground-truth points $(x_{i,t},y_{i,t})$ for which $(\hat{x}_{i,t},\hat{y}_{i,t})$ is predicted as unoccluded, and the spatial prediction is within a distance of $\alpha$.
Let $FN_{\alpha}$ be false negatives: visible ground truth points which are predicted to be occluded, or for which the predicted spatial position is farther than $\alpha$ from the ground truth.  Let $FP_{\alpha}$ be false positives: points $(\hat{x}_{i,t},\hat{y}_{i,t})$ which are predicted to be visible, where the ground truth is farther than distance $\alpha$ or where the ground truth is marked as occluded.
The Jaccard metric is then $|TP_{\alpha}|/(|TP_{\alpha}|+|FN_{\alpha}|+|FP_{\alpha}|)$.  In practice, we compute these metrics across 5 different thresholds of the form $\alpha=\delta^{x}$ pixels, for $x \in \{0,1,2,3,4\}$ and $\delta=2$.  We compute an average across thresholds to get an overall metric. 

\paragraph{Baselines}
We next define a baseline method that can begin to solve the point tracking problem. 
One of the closest problems in the literature is segment tracking, using datasets like DAVIS~\cite{pont20172017}.  Therefore, our baseline is inspired by VFS~\cite{xu2021rethinking}, a state-of-the-art method for DAVIS.
VFS has two key components: first, a self-supervised training phase where the aim is to learn a good similarity metric between points across images, and second, a test-time tracking algorithm based on earlier work~\cite{wang2019learning} which can associate points in unlabeled frames with the points in labeled frames.
Our model, however, is modified to deal with points rather than segments, and to leverage the labeled training data we have available.
For pre-training, we adopt a contrastive approach~\cite{henaff2020data}.
We use a standard ResNet-50~\cite{he2016deep} as a backbone, up to the final convolution, and with $\mbox{stride}=1$ for the final two blocks, which gives us a feature grid $F_{i}$ (which is $L2$-normalized over channel axis) for each frame, at stride $8$.  Given a query point, $(x^{*},y^{*},t^{*})$~(note: we drop the $i$ index for clarity as we consider a single point), we first extract a feature $f^{*}$ for that point, via bilinear interpolation at position $(x^{*}/8,y^{*}/8)$ from the feature grid for frame $t^{*}$.
We then compute the following contrastive loss function:
\begin{equation}
    \sum_{f_j\in F_{i}}\gamma_{j} \log\left(\frac{\exp(f^{*}_i\cdot f_j)/\tau}{\sum_{k} \exp(f^{*}_i\cdot f_k/\tau)}\right)
\end{equation}
where $j$ and $k$ index over the spatio-temporal dimensions of $F$.
The temperature hyperparameter $\tau$ is set to $0.1$.
$\gamma_{j}$ is the source of supervision: its value is high if $f_j$ is in correspondence with $f^{*}$, and it is 0 if they aren't.
Note that if there is exactly one other point considered to be ``in correspondence'' with $f^{*}$, then the sum over $k$ has a single term, and we are left with a standard contrastive loss.
However, in our case, we have multiple potential correspondences, and this loss encourages all of them to have roughly equally high dot products.
 
We compute $\gamma_{j}$ via bilinear interpolation.  Say that the feature $f_j$ is on frame $t$ at position $\hat{x},\hat{y}$ within the convolutional feature grid, and the ground truth position is at $(x_t/8,y_t/8)$~(in the coordinate frame of the feature grid).
If $(x_t/8,y_t/8)$ lies within the grid cell which has one of its corners at $\tilde{x_j},\tilde{y_j}$, then we set $\gamma_{j}=(1-|\tilde{x_k}-x_t/8|)*(1-|\tilde{y_k}-y_t/8|)$.  Otherwise, it is 0.  $\gamma_{j}$ is also set to 0 if the ground truth is marked as occluded for frame $t$.

At test time, given a query point $(x^{*},y^{*},t^{*})$, we begin by computing a correspondence to every frame.  For a single frame at time $t$, this is done via a dot product between $f^{*}$ (again extracted via bilinear interpolation) and each feature in frame $t$, followed by a softmax $S$ across space, which gives us a heatmap of likely locations, i.e., $S_{xy}$ is the probability that $x,y$ corresponds to $(x^{*},y^{*},t^{*})$.  We then compute $\tilde{S}$ by finding the argmax of $S$ and zeroing out any grid cells further than 40 pixels away (5 grid units) from it, to suppress potential multimodal correspondences.  Then we compute the weighted average of the potential locations $[\hat{x},\hat{y}]=\sum_{x,y}[x,y]*\tilde{S}_{xy}/\sum_{x,y}\tilde{S}_{xy}$.  

In order to classify whether the point is occluded, we use cycle consistency~\cite{wang2019learning,wang2019web}.  That is, we extract a new feature $\hat{f}$ from point $[\hat{x},\hat{y}]$ in frame $t$, and reverse the process, computing a softmax over locations in frame $t^{*}$ and converting it into a point correspondence.
If the estimated point is further than $48$ pixels of its starting location, we consider it occluded.

We evaluate this procedure on MOVi-E at a resolution of $256\times 256$.  For each evaluation video, we sample 256 query points randomly across all frames.  We attempt to sample an equal number of points from each object as well as the background, but cap the number of samples per object at a maximum of 0.2\% of the visible pixels.  We use standard data augmentations, including a random crop of the image as small as 30\% of the image area and an aspect ratio between 2:1 and 1:2, and the same color augmentations and pixel noise that was used for our optical flow experiments.

Results are shown in~Table~\ref{tab:point_tracking}.
For comparison, we also include a na{\"i}ve baseline which assumes no motion and no occlusion, which is the best that can be done without reference to the pixels.
We see that the contrastive approach is fairly good at coarse tracking, reducing the error rate on the largest threshold from 25.2\% down to just 5.8\%, a reduction of more than 6 times relative to the na{\"i}ve baseline.  However, for more precise tracking, the reduction in error is not nearly as great.  On the other hand, accuracy at detecting occlusions is quite poor for this method; for the threshold we used for cycle consistency (48), the accuracy is actually worse than chance.  However, points for which the cycle consistency check failed are actually unlikely to be within any distance threshold; therefore, we find that removing these points from the output improves the average Jaccard metric.

Because the network is trained and evaluated on data from the same distribution, there is a possibility that the algorithm is memorizing some aspects of the training data, such as the common trajectories followed by objects.  To evaluate out-of-distribution transfer, we also applied our algorithm to vertically-flipped videos, and show the results in Table~\ref{tab:point_tracking_vflip}.  This harms performance by about 4\%, suggesting that the network is memorizing trajectories to a small extent.

Finally, Figure~\ref{fig:point_tracking} shows a qualitative example of our point tracking algorithm on a kubric validation video.  We see that for easy points with relatively little motion, like the blue star or cyan circle, the algorithm is quite accurate, even when there is relatively little texture.  Tracking can also be good for the points that have reasonably distinctive texture, like the green square.  Occlusions are also sometimes detected correctly: for the red X, the algorithm correctly determines the occlusion on frame 11, although it prematurely finds the point again on frame 15.  However, there are also obvious failures: for instance, the algorithm loses the magenta plus, likely because the object has rotated so much that the appearance has changed substantially.  This suggests that it may be useful to employ global reasoning about the orientation of objects rather than relying on appearance alone.  We also see a very large error for the diamond on frame 7, when the point is occluded; the algorithm instead places the point between the occluder objects, possibly because the feature descriptors are relying too heavily on context.  This result suggests that simultaneously capturing global object motion while remaining robust to occlusion will be a substantial challenge in this dataset moving forward. 
\newcommand{\SceneSemanticSegmentation}{Scene Semantic Segmentation\xspace}
\newcommand{\Ray}{\mathbf{r}}
\newcommand{\RayOrigin}{\mathbf{o}}
\newcommand{\RayDirection}{\mathbf{d}}
\newcommand{\DeepLab}{DeepLab\xspace}
\newcommand{\SparseConvNet}{SparseConvNet\xspace}
\newcommand{\NeSF}{NeSF\xspace}
\newcommand{\Klevr}{KLEVR\xspace}
\newcommand{\ToyBox}{ToyBox\xspace}
\newcommand{\ToyBoxEasy}{ToyBox5\xspace}
\newcommand{\ToyBoxHard}{ToyBox13\xspace}


\begin{figure}[t]
\centering
\small
\begin{tabular}{@{}llccc@{}}
    \toprule
    & & \Klevr & \ToyBoxEasy & \ToyBoxHard \\
    \midrule
    \multirow{2}{*}{2D}
    & \DeepLab~\cite{chen2017deeplab} & 97.1\% & 81.6\% & 63.1\% \\       
    & \NeSF~\cite{vora2021nesf}       & 92.7\% & 81.9\% & 56.5\% \\       
    \midrule
    \multirow{2}{*}{3D}
    & \SparseConvNet~\cite{graham2017submanifold} & 99.7\% & 93.4\% & 83.2\% \\    
    & \NeSF~\cite{vora2021nesf}                   & 97.8\% & 88.7\% & 60.1\% \\  
    \bottomrule
\end{tabular}
\vspace{1em}
\captionof{table}{
    \textbf{\SceneSemanticSegmentation} -- 
    We compare mean intersection-over-union in 2D image segmentation (top) and 3D point cloud segmentation (bottom) on the three datasets.
}
\label{tab:toybox_results}

\vspace{2em}
\newcommand{\minifigure}[2]{\includegraphics[width=6em]{#2}}
\begin{tabular}{ccc}
\Klevr
    & \ToyBoxEasy
    & \ToyBoxHard
    \\
\minifigure{\ToyBoxEasy}{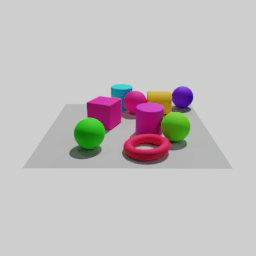}
  & \minifigure{\ToyBoxEasy}{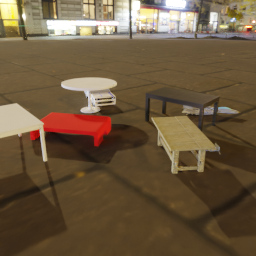}
  & \minifigure{\ToyBoxHard}{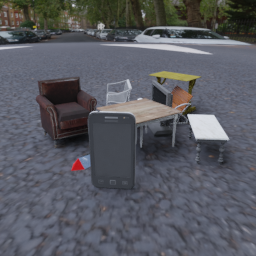}
  \\
\minifigure{\ToyBoxEasy}{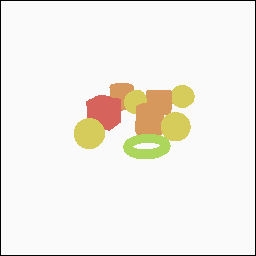}
  & \minifigure{}{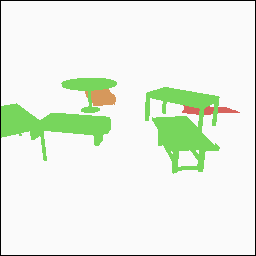}
  & \minifigure{}{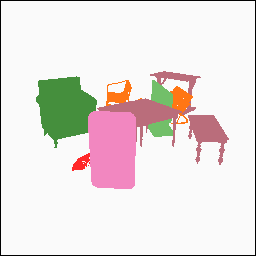}
  \\
\end{tabular}
\vspace{1em}
\captionof{figure}{
    \textbf{\SceneSemanticSegmentation} -- 
    Example RGB and segmentation renders from the datasets.
}
\label{fig:toybox_example}
\end{figure}

\subsection{\SceneSemanticSegmentation}
\label{sec:toybox}

As another use case, we consider the task of comparing 2D and 3D semantic segmentation models.
As these methods operate on fundamentally different substrates, it is challenging to quantify the effectiveness of one method versus another.
To this end, we construct three synthetic datasets where 2D image and 3D point cloud are in direct correspondence: \Klevr, \ToyBoxEasy and \ToyBoxHard.
Each of the \ToyBox datasets each consists of 525 scenes.
Each scene contains 4-12 upright ShapeNet objects on a flat surface with one of 382 randomly chosen HDRI backdrops.
The datasets differ only in the set of ShapeNet objects employed.
In \ToyBoxEasy, we use the top 5 most common object categories; in \ToyBoxHard, the 13 most common object categories.
For each scene, we select 300 camera poses and render 3 images per pose: an RGB map, a segmentation map, and a depth map.
With knowledge of camera parameters, we are able to construct a camera ray $\Ray(t) = \RayOrigin + t \RayDirection$ corresponding to each pixel in the dataset.
When combined with depth and segmentation maps, we obtain a labeled 3D point cloud where each 3D point corresponds to one camera pixel.
We define a scene's point cloud to be the union of all 3D points constructed from all camera poses.
The \Klevr dataset is constructed identically to the \ToyBox datasets except with a fixed, neutral-grey backdrop and 5 platonic object shapes.

\paragraph{Experiments}
We demonstrate two representative baselines for 2D image and 3D point cloud segmentation: \DeepLab~\cite{chen2017deeplab} and \SparseConvNet~\cite{graham2017submanifold}, respectively.
In addition, we compare these methods with \NeSF~\cite{vora2021nesf}, a method for dense 2D and 3D scene segmentation from posed RGB images.  
We train all methods with semantic supervision derived from 9 cameras per scene from 500 scenes and hold out 4 cameras per scene from the remaining 25 scenes for evaluation.
In 2D, we see that \NeSF and \DeepLab perform similarly, with \DeepLab outperforming \NeSF by 0.3\% to 6.6\% (\Table{toybox_results}).
In qualitative results, we find that \DeepLab's predictions more tightly outline objects at the expense of multiview consistency.
In 3D, we see that \SparseConvNet achieves between 1.9\% and 23.1\% higher mean intersection-over-union than \NeSF with larger margins as dataset complexity increases.
We attribute this to the \SparseConvNet's access to ground truth 3D geometry and supervision in the form of a dense point cloud.
This results in an exceedingly dense and accurate representation of 3D geometry.
\NeSF, on the other hand, must infer 3D geometry and semantics from posed 2D images alone.
Further results and comparison to NeSF are presented in \cite{vora2021nesf}. 
\newcommand{\NVS}{Conditional Novel View Synthesis\xspace}
\newcommand\mspic[1]{
    \includegraphics[height=0.105\linewidth]{fig/nvs/#1}
}
\begin{figure}[t]
\small
\centering 
\begin{tabular}{@{}lccc}
\toprule
 & PSNR & SSIM & LPIPS \\
\midrule
LFN \cite{sitzmann2021lfns} & 14.77 & 0.328 & 0.582 \\
PixelNeRF \cite{yu2021pixelnerf} & 21.97 & 0.689 & 0.332 \\       
SRT \cite{srt_arxiv} & 23.41 & 0.697 & 0.369 \\       
\bottomrule
\end{tabular}
\vspace{1em}
\captionof{table}{
\textbf{\NVS} --
Quantitative evaluation for novel view synthesis.
}
\label{tab:template}

\setlength{\tabcolsep}{0mm}
\def\arraystretch{1}
\begin{tabular}{c@{\hskip -.5mm}c@{\hskip -.5mm}c@{\hskip -.5mm}c@{\hskip -.5mm}ccccc}
    \multicolumn{5}{c}{Input Images} &
    \cite{sitzmann2021lfns} &
    \cite{yu2021pixelnerf} &
    \cite{srt_arxiv} &
    Target
    \\
    \mspic{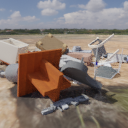} &
    \mspic{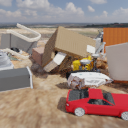} &
    \mspic{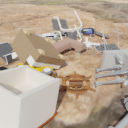} &
    \mspic{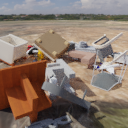} &
    \mspic{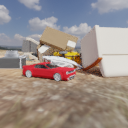} &
    \mspic{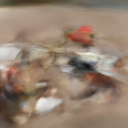} &
    \mspic{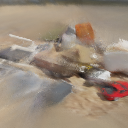} &
    \mspic{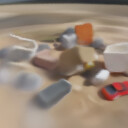} &
    \mspic{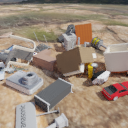}
    \\
    \mspic{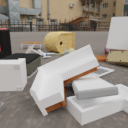} &
    \mspic{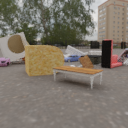} &
    \mspic{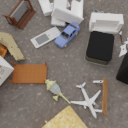} &
    \mspic{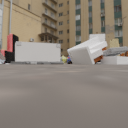} &
    \mspic{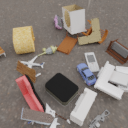} &
    \mspic{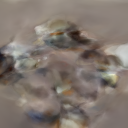} &
    \mspic{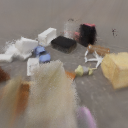} &
    \mspic{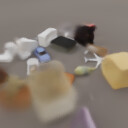} &
    \mspic{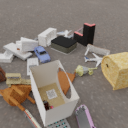}
    \\
    \mspic{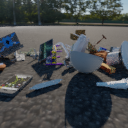} &
    \mspic{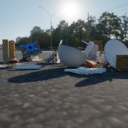} &
    \mspic{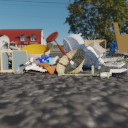} &
    \mspic{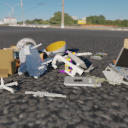} &
    \mspic{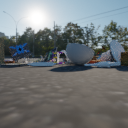} &
    \mspic{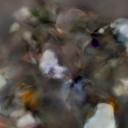} &
    \mspic{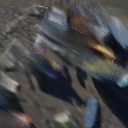} &
    \mspic{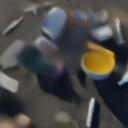} &
    \mspic{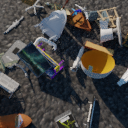}
    \\[4mm]
\end{tabular} 
\captionof{figure}{
    \textbf{\NVS} --
    The scenes are very difficult as they contain a large number of objects in random poses, contain realistic backgrounds, and are rendered with ray-tracing.
    LFN fails to capture these datasets in a global latent, severely under-fitting.
    PixelNeRF shows better quality renders, which however degrade for out-of-training distribution target views(\eg, bottom row).
} 
\label{fig:nvs}
\end{figure}

\subsection{\NVS}
\label{sec:nvs}
%
Neural scene representations such as~\cite{nerf} have led to state-of-the-art results in novel-view synthesis tasks on real-world datasets~\cite{nerfw}, however, a new model must be trained on each new scene, which does not allow the model to learn a \textit{prior} over the dataset.
Prior works commonly use the ShapeNet dataset~\cite{chang2015shapenet} to evaluate how well a method can generalize to novel scenes~\cite{yu2021pixelnerf, sitzmann2021lfns}.
However, ShapeNet consists of \textit{canonically} oriented objects with thousands of same-class examples for some categories (\eg, airplanes) rendered with flat shading.
We introduce a new large-scale dataset using Kubric to generate photo-realistic scenes with groups of ShapeNet objects.
Each of the 1M scenes use one of 382 randomly chosen background maps.
We render ten $128\times128$ random views for each scene and use five as \emph{conditioning views}, and the other five as \emph{target views}.
The task is to reconstruct the target views \textit{given} the conditioning views.
We compare the following recent methods: PixelNeRF~\cite{yu2021pixelnerf}, which projects points into the conditioning views to interpolate encoded image features, LFN~\cite{sitzmann2021lfns}, which condenses the scene into a single latent code, and decodes it into an implicit scene representation using a hyper-network that produces the weights of the scene-specific MLP, and SRT~\cite{srt_arxiv}, which learns a scalable set-latent scene representation through a transformer encoder-decoder architecture.
\cref{fig:nvs} compares these methods on our new challenge.
While all methods above produce fairly high-quality results on ShapeNet~(see~\cite{srt_arxiv}), they scale vastly differently to our photorealistic, complex dataset: while PixelNeRF has apparent difficulties with views that are far away from the conditioning views, LFN does not scale at all to this complexity, and SRT's reconstructions suffer from blurriness.
Further details on this challenge and the dataset release are presented in \cite{srt_arxiv}.
 
\section{Conclusions}

We introduce Kubric, a general Python framework complete with tools for generation at scale, integrating assets from multiple sources, rich annotations and a common export data format (SunDS) for porting data directly into training pipelines.
Kubric enables the generation of
high quality synthetic data, addressing many of the  problems inherent in curating natural image data, and circumventing the expense of building task-specific, one-off pipelines.
We demonstrate the effectiveness of our framework in 11 case studies with generated datasets of varying complexity for a range of different vision tasks. 
In each case, Kubric has substantially reduced the engineering effort to generate the required data and has facilitated reuse and collaboration.
We hope that it will help the community by lowering the barriers to generating high-quality synthetic data, reduce fragmentation, and facilitate the sharing of pipelines and datasets.

\paragraph{Limitations and future work} 
While already tremendously useful, Kubric is still a work in progress and does not yet support many   features of Blender and PyBullet.
Notable examples include volumetric effects like fog or fire, soft-body and cloth simulations, and advanced camera effects such as depth of field and motion blur.  
We also plan to preprocess and unify assets from more sources, including the ABC dataset~\cite{koch2019abc} or Amazon Berkeley Objects~\cite{collins2021abo}.
At present, Kubric requires substantial computational resources due to its reliance on a path-tracing renderer versus a rasterizing renderer.
We hope to add support for a rasterizing backend, allowing users to trade-off speed and render quality.


We include a discussion on the potential societal impact and ethical implications surrounding the application of our system in Section~\ref{sec:ethics} of the supplementary material.
\section{Societal Impact and Ethical Considerations}
\label{sec:ethics}
The diversity of applications that can leverage our system merits a broad discussion on both its potential societal impact and the ethical considerations one should consider when applying it. 
As a general purpose tool rather than a application-targeted solution, Kubric bears the potential for diverse benefits as well as the risk of harm through negligence or even malicious misuse. 

\paragraph{Societal Impact} 
Synthetic dataset construction presents system engineers with the opportunity to detect and correct potentially dangerous failure modes -- particularly for those applications that involve human interaction, e.g., self-driving cars and robotics -- \textit{prior to} their deployment in the wild and with real humans.
This does not, however, eliminate the possibility of developing systems that could cause serious injury to humans; instead, it raises the importance of considering the possibility of such outcomes.
We therefore urge Kubric users that work towards systems deployed outside the laboratory, to take \textit{active} measures towards mitigating such risks and consider them at the forefront of the design process. 
Still, the potential value of leveraging synthetic dataset construction as a tool to help guard against harmful outcomes should be further explored.

\paragraph{Ethical Considerations} 
While Kubric's dataset generation relies on human-driven design automation, it sidesteps the immediate need for human-derived data.
This can help to avoid ethical problems and legal obstacles to research and can also be a powerful tool for studying and mitigating undesirable societal biases.
Still, any human-in-the-loop semi-automated processes are susceptible to the biases of their designers.
While a more explicitly-controlled dataset design methodology allows engineers to postpone complications surrounding the (important) privacy concerns due to the treatment of data captured from the real-world, one can reasonably argue that such a benefit is offset -- at least in part -- by biases introduced during the dataset synthesis process.
Indeed, any explicitly-constructed synthetic dataset will be vulnerable to inheriting the biases of the processes employed when constructing it -- but we argue that it promotes the discussion and (hopefully) the mitigation of biases at \textit{both} an earlier stage of the design process \textit{and} with a greater degree of controllability.
The potential downstream impacts of distributional differences between the synthetic and real-world data would then become an important additional concern, requiring explicit evaluation and potential mitigation/treatment to safeguard against real world bias.

We also note that, while Kubric, on the one hand, provides an effective way to create new datasets, helping to avoid becoming stuck on, and over-fitting to existing data, it may also, on the other hand, enable the proliferation of datasets tailored to highlight the advantages of one's method of choice.
While this is true with all dataset creation, it is hoped that through experimentation and replication, as with model architectures, the field will self-select datasets that are useful, providing fair, balanced assessment of different models on tasks of common interest.

\paragraph{Environmental considerations}
Controllable synthetic dataset construction helps to promoting a control-based scientific methodology: e.g., where confounding factors can be explicitly isolated and tested against, and by allowing smaller problems to be constructed (i.e., where only those behaviors one seeks to validate are tested against) before scaling to larger, more general settings.
This strategy can help to reduce the need for repeatedly training large-scale models on huge datasets, and thus lower the overall environmental impact of the research project.
However, the ability of to generate large and controllable synthetic datasets does not come without its costs; for example, our optical flow dataset required roughly 3 CPU-years of compute-time.
Hence we urge researchers to be mindful of the costs of both training and generating datasets, and to avoid generating unnecessarily large datasets. 
As such, the design of surrogate synthesis models capable of augmenting synthetic datasets in a more energy-efficient fashion, e.g., with latent-space dynamical models, is an important line of future research.



\paragraph{Synopsis}
Synthetic data offers the opportunity for researchers and engineers to consider and face the impact of bias on their systems, to design around detecting dangerous failure modes, and all in a privacy-preserving setting.
This certainly does not preclude the presence of such issues in any resulting final system, as the manner and degree in which these opportunities are leveraged should mandate an additional level of responsibility during the design and meta-evaluation of the synthetic datasets.

{
    \small
    \bibliographystyle{ieee_fullname}
    \bibliography{macros,main}
}

\end{document}